\documentclass[final,3p,times,authoryear]{elsarticle}

\usepackage{amssymb}

\usepackage{amsmath}
\usepackage{tikz}
\usetikzlibrary{calc}
\usepackage{graphicx}      
\usepackage{natbib}        
\usepackage{pifont}
\usepackage{tabularx}
\usepackage{booktabs}
\usepackage{amsthm}
\usepackage{amsmath}
\usepackage{multirow}
\usepackage{graphicx}
\usepackage{longtable}
\usepackage{csvsimple}
\usepackage{adjustbox}
\usepackage{makecell}

\usepackage{siunitx}
\usepackage[ruled,vlined]{algorithm2e}
\usetikzlibrary{positioning}
\usepackage{listings}
\usepackage{xcolor}
\usepackage{url} 
\usepackage{float} 
\usepackage{xurl}
\usepackage{graphicx}
\usepackage{subcaption}
\usepackage{multicol}
\usepackage{microtype}

\usepackage{array}
\usepackage{ragged2e}
\usepackage{caption}
\usepackage{environ}

\usepackage[hidelinks]{hyperref}
\usepackage{orcidlink}

\lstdefinelanguage{yaml}{
  keywords={true,false,null,y,n},
  keywordstyle=\bfseries,
  morekeywords=[2]{LayoutDomain,ArticlesDomain,OrdersDomain,%
    ResourcesDomain,ResourceDomain,StorageDomain,requirements,constraints},
  keywordstyle=[2]\color{casopDark}\bfseries,
  basicstyle=\ttfamily\small,
  sensitive=false,
  comment=[l]{\#},
  commentstyle=\color{gray}\ttfamily,
  morestring=[b]',
  morestring=[b]",
  literate =    {---}{{\ProcessThreeDashes}}3
                {\ -\ }{{\mdseries\ -\ }}3,
}

\lstdefinestyle{cardstyle}{
  language=yaml,
  basicstyle=\scriptsize\ttfamily,
  breaklines=true,
  breakatwhitespace=false,
  frame=lines,
  framesep=1mm,
  rulecolor=\color{gray!40},
  keepspaces=true,
  columns=fullflexible,
  aboveskip=0pt, belowskip=0pt,
}

\usepackage[capitalise]{cleveref}
\crefname{equation}{Equation}{Equations}
\Crefname{equation}{Equation}{Equations}
\crefname{figure}{Figure}{Figures}
\Crefname{figure}{Figure}{Figures}
\crefname{table}{Table}{Tables}
\Crefname{table}{Table}{Tables}

\newcommand{\frameworkname}{CASOP}
\newcommand{\frameworklong}{Context-Aware Synthesis of Optimization Pipelines}

\newcommand{\nsets}{7}

\newcommand{\npipelines}{1,063,044}

\definecolor{casopDark}{HTML}{073B53}
\definecolor{casopBlue}{HTML}{3EB7E6}
\definecolor{casopLightBlue}{HTML}{BFEAF6}

\newcommand{\algstar}{\textsuperscript{*}}
\newcommand{\algindent}{\hspace{1.2em}}
\newcommand{\algtablefont}{\footnotesize}

\NewEnviron{algorithmtable}[3][htbp]{%
  \begin{table}[#1]
  \centering
  \caption{#2}
  \label{#3}
  \begingroup
  \algtablefont
  \setlength{\tabcolsep}{4pt}
  
  \begin{tabularx}{\linewidth}{@{}
    >{\RaggedRight\arraybackslash}p{0.37\linewidth}
    >{\Centering\arraybackslash}p{0.07\linewidth}
    >{\RaggedRight\arraybackslash}X
  @{}}
  \BODY
  \end{tabularx}
  \endgroup
  \end{table}
}

\NewEnviron{configtable}[3][htbp]{%
  \begin{table}[#1]
  \centering
  \caption{#2}
  \label{#3}
  \begingroup
  \algtablefont
  \setlength{\tabcolsep}{4pt}
  
  \begin{tabularx}{\linewidth}{@{}
    >{\RaggedRight\arraybackslash}p{0.24\linewidth}
    >{\RaggedRight\arraybackslash}p{0.31\linewidth}
    >{\RaggedRight\arraybackslash}X
  @{}}
  \BODY
  \end{tabularx}
  \endgroup
  \end{table}
}

\journal{Journal to be define}

\begin{document}

\begin{frontmatter}

\title{Context-Aware Synthesis of Optimization Pipelines for Warehouse Optimization}

\author[inst1]{Janik Bischoff\,\orcidlink{0009-0007-6592-9768}}
\author[inst1]{Anne Meyer\,\orcidlink{0000-0001-6380-1348}}
\author[inst2]{Uta Mohring\,\orcidlink{0000-0001-9218-0536}}
\author[inst3]{Fabian Dunke\,\orcidlink{0000-0002-4805-9576}}
\author[inst4]{Maximilian Barlang\,\orcidlink{0009-0005-8081-4286}}
\author[inst3]{Özge Nur Subas}
\author[inst1]{Hadi Kutabi\,\orcidlink{0000-0002-6023-7742}}
\author[inst3]{Stefan Nickel\,\orcidlink{0000-0002-8339-0117}}
\author[inst4]{Kai Furmans\,\orcidlink{0000-0001-6009-5564}}

\affiliation[inst1]{organization={Institute for Information Management in Engineering, Karlsruhe Institute of Technology},
            country={Germany}
            }
\affiliation[inst2]{organization={Department of Business Administration, University of Zurich},
            country={Switzerland}}
            
\affiliation[inst3]{organization={Institute for Operations Research, Karlsruhe Institute of Technology},
            country={Germany}
            }

\affiliation[inst4]{organization={Institute for Material Handling and Logistics,Karlsruhe Institute of Technology},
            country={Germany}
            }

\begin{abstract}

Order fulfillment in manual picker-to-goods warehouses involves interconnected decisions such as item assignment, order batching, and picker routing. 
While integrated models capture interactions between these decisions, practical warehouse systems often require decomposed approaches due to organizational boundaries, differing responsibilities, or limited data availability. 
Existing studies primarily evaluate algorithms for isolated subproblems or fixed subproblem combinations for specific warehouse settings, but lack a general mechanism to determine applicable algorithm configurations, compose them into valid solution pipelines, and assess their performance.

With \frameworklong{} (CASOP), we propose a framework for constructing and evaluating context-specific optimization pipelines and apply these to order fulfillment.
The framework comprises: (1) a modular repository of algorithms for common order fulfillment problems; (2) semantic data and algorithm cards to describe warehouse context and algorithm requirements; (3) a taxonomy that structures order fulfillment problems into relevant subproblems; (4) a pipeline synthesizer that identifies applicable algorithms for a given warehouse context and composes all valid optimization pipelines; and (5) a pipeline evaluator that assesses all resulting pipelines.
We demonstrate the framework on \nsets{} benchmark instance sets covering four problem classes, resulting in \npipelines{} valid pipelines.  
The framework supports researchers and practitioners in designing, automatically synthesizing, and selecting valid, high-performing algorithmic pipelines for warehouse operations. The software is open-source and available at \href{https://github.com/kit-dsm/ware_ops_pipes}{github.com/kit-dsm/ware\_ops\_pipes} and \href{https://github.com/kit-dsm/ware_ops_algos}{github.com/kit-dsm/ware\_ops\_algos}.

\end{abstract}





\end{frontmatter}



\section{Introduction}
\label{sec:introduction}
\begin{figure}[ht]
    \centering
    \includegraphics[width=1\linewidth]{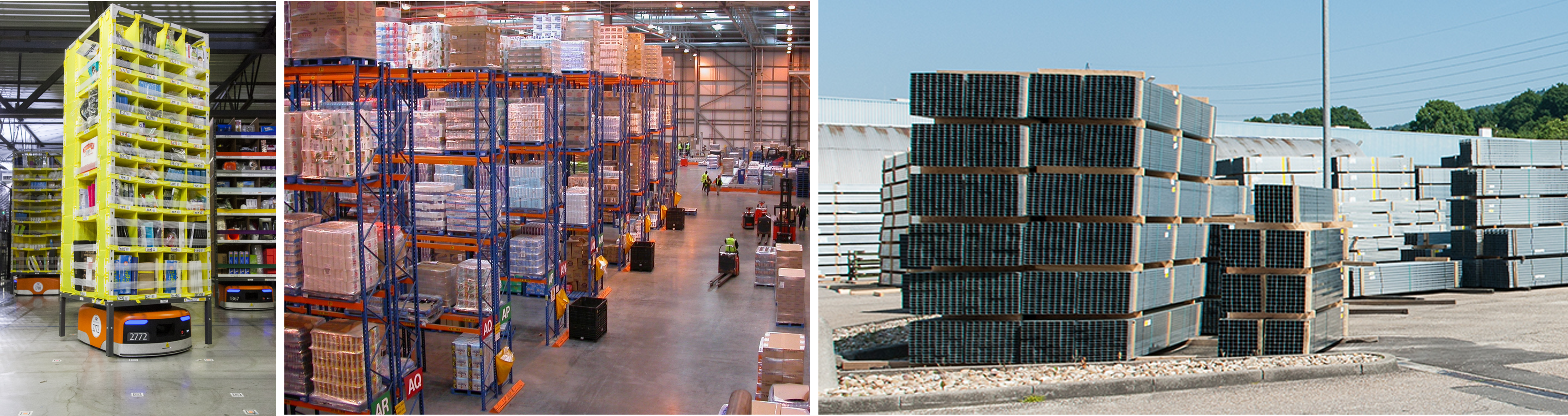}
    \caption{Overview of different types of warehouse systems. A robotic mobile fulfillment center of Amazon (left, \protect\citet{amazon_fulfillment_press}, reproduced with permission). The distribution center of a grocery retailer (center, photo by Nick Saltmarsh, CC BY 2.0, via Wikimedia Commons). A block-stacking warehouse for metal profiles (right, courtesy of Protektorwerk Florenz Maisch GmbH \& Co. KG, reproduced with permission).}
    \label{fig:warehouse_systems}
\end{figure}

Warehouse systems differ widely in their physical layout, technical equipment, IT systems, degree of automation, and operating processes. Figure~\ref{fig:warehouse_systems} illustrates this variation with examples ranging from robotic fulfillment systems over manual distribution centers to block-stacking warehouses. 
These differences affect not only operational performance, but also which optimization problems arise and which solution approaches can be applied.

In the broader context of warehouse optimization, this paper focuses on order fulfillment in manual picker-to-goods systems, where pickers walk to storage locations to retrieve items ordered by customers, typically using carts to carry the retrieved items.
Unlike automated warehouse systems, whose fixed hardware capacity is typically dimensioned for peak loads and may therefore remain underutilized during periods of lower demand, manual picker-to-goods systems retain substantial operational flexibility \citep{boysen50years}.

However, this flexibility can only be effectively exploited if the arising operational decision problems are addressed.
Order fulfillment comprises several interrelated decision problems: item assignment, order batching, picker routing, and picker scheduling. 
The \textit{item assignment problem} selects, for each requested article, the storage location(s) from which it is picked. 
The \textit{order batching problem} groups customer orders into sets retrieved in a single picker tour, subject to picker capacity. 
The \textit{picker routing problem} determines the path for collecting items from a given pick list, typically modeled as a variant of the traveling salesperson problem (TSP). 
The \textit{picker scheduling problem} assigns batches or tours to pickers and orders their execution, with objectives such as tardiness or makespan.

Each subproblem has been extensively studied in isolation and partially integrated approaches have also been studied, combining, for example, batching and routing; we refer to \citep{de_koster_design_2007,gu_research_2007,BOYSEN2019396,pardo_order_2024,bock2025survey} for comprehensive reviews.
Although integrated solution approaches for subsets of these problems in order fulfillment exist (see e.g. \citet{van_gils_formulating_2019}), their applicability in practical settings is limited due to organizational boundaries, heterogeneous scopes of responsibility, or limited data availability \citep{boysen50years}. 
Motivated by this, \citet{boysen50years} recently raised the following important research question: which planning problems in warehouse operations can be decomposed into independently solved models while still preserving substantial solution quality relative to a comprehensive model?
This shifts the focus from proposing ever-larger integrated models to understanding which decompositions are valid, practically applicable, and well-performing in a given warehouse context.

Existing work provides only partial answers to this question.
For example, \citet{pardo_order_2024} propose a taxonomy to structure variants of order batching problems into subproblems.
However, the taxonomy does not cover the full order fulfillment process considered here.
In particular, it does not consider item assignment, treats routing as a batching-related task, and does not represent picker scheduling as a separate subproblem.
In general, benchmark studies from the warehouse literature often compare a limited set of manually selected algorithms for a specific problem setting \citep{petersen_ii_evaluation_1999,van_gils_formulating_2019,bahceci_evaluation_2022}.
This becomes limiting when a larger number of order fulfillment subproblems and possible configurations of individual algorithms are considered simultaneously.
A solution approach may solve the subproblems sequentially, or it may solve selected parts in an integrated way, for example, batching and routing. The algorithms themselves might be configurable, and different configurations might perform better or worse depending on the warehouse context. 
Furthermore, not every algorithm is applicable to a given warehouse context, and not every combination of algorithms forms a valid solution approach.
Algorithms may exploit specific layout structures such as those of parallel-aisle warehouse layouts \citep{hesler_exact_2024}. Other solution approaches for subproblems, such as item assignment, are only required if the warehouse follows a scattered storage policy.  

Therefore, answering the decomposition question raised by \citet{boysen50years} requires a framework that is able to not only select the best algorithm based on performance,
but also identify applicable algorithms, compose them into valid optimization pipelines, and evaluate them in a given warehouse context.

This connects warehouse optimization to three related research streams.
Algorithm selection provides methods for choosing among alternative solvers for a given problem instance \citep{xu2008satzilla,bischl2016aslib}.
Semantic modeling provides means to describe domain knowledge and make assumptions explicit \citep{negri_modelling_2017,knoll2019developing}.
Type-based synthesis approaches, such as the Combinatory Logic Synthesizer, show how repositories of typed components can be used to generate valid compositions, for example, for scheduling heuristics \citep{maeckel_scheduling_cls}.
CLS-Luigi extends this idea to the generation and execution of executable data pipelines from algorithm repositories \citep{meyer_cls-luigi_2024}.
However, their integration for complex warehouse optimization remains largely unexplored.
None of the existing frameworks links warehouse context, algorithm requirements, pipeline construction, and performance evaluation across multiple order fulfillment problems.

\begin{figure}[H]
    \centering
    \includegraphics[width=0.7\linewidth]{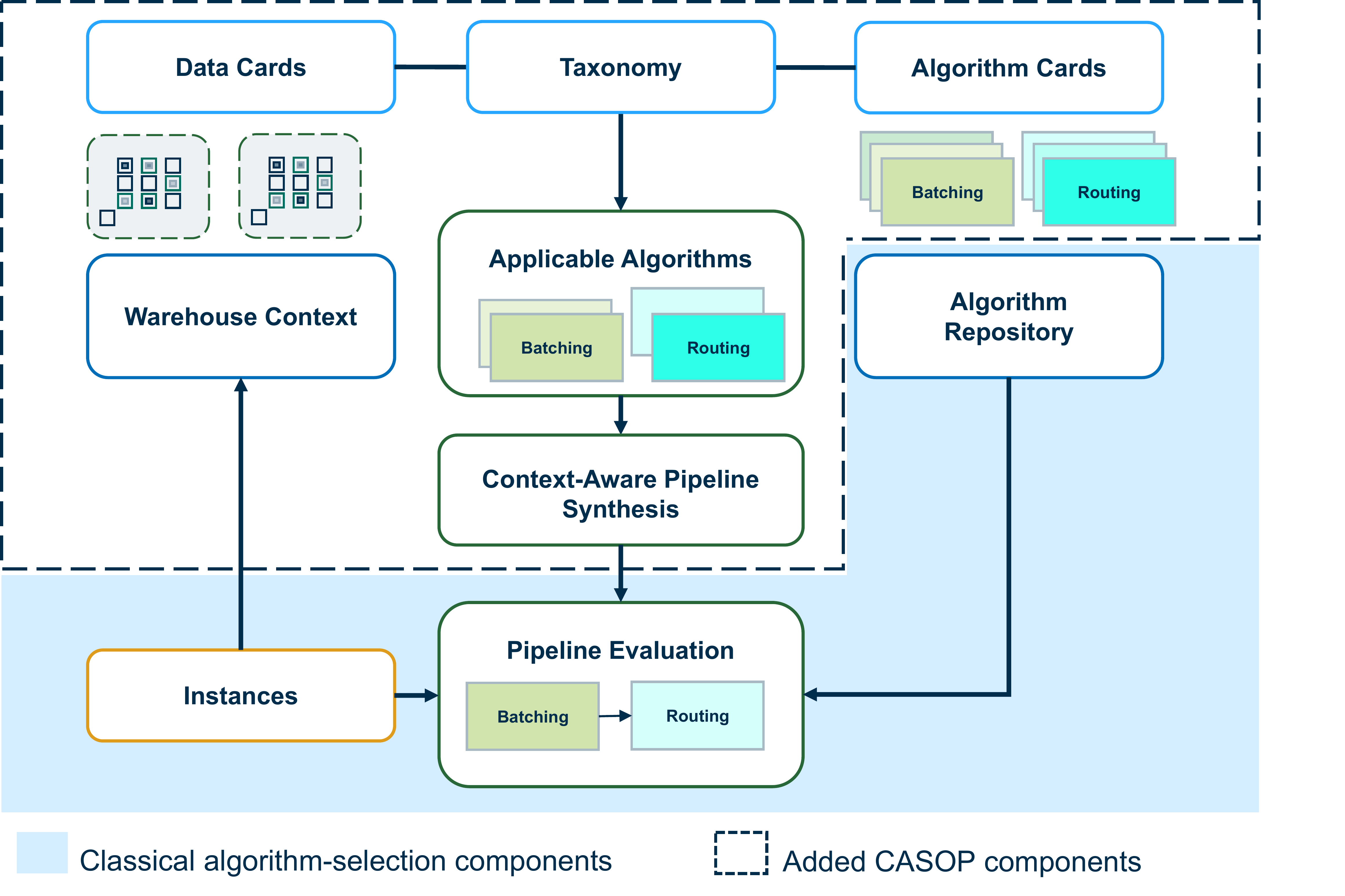}
    \caption{Main contributions of the paper and delimitation from classical algorithm selection. Unlike classical algorithm selection, CASOP explicitly considers the warehouse context and provides automated pipeline synthesis.}
    \label{fig:visual-abstract}
\end{figure}

We propose CASOP, a framework for context-aware synthesis of optimization pipelines for warehouse optimization. As we illustrate in Figure \ref{fig:visual-abstract}, we use a taxonomy to decompose problems into their subproblems and go beyond traditional algorithm selection by explicitly considering warehouse context and algorithm requirements in the form of semantic data and algorithm cards.
This allows us to compose and identify well-performing and valid optimization pipelines from an algorithm repository tailored to a warehouse context.
CASOP comprises the following features:

\begin{enumerate}
    \item \textbf{A modular algorithm repository}: Reusable algorithm implementations for key order fulfillment subproblems provided through harmonized interfaces.
    
    \item \textbf{Semantic representation of warehouse context and algorithm requirements}: A machine-readable description of warehouse system characteristics in the form of a data card, covering layout, articles, orders, resources, and storage. 
    Machine-readable description of algorithm requirements in the form of an algorithm card.

    \item \textbf{Problem taxonomy and pipeline template}: A taxonomy decomposing order fulfillment problems into subproblems and a corresponding pipeline template covering four problem classes.
    
    \item \textbf{Context-specific algorithm selection}:  
    Matching of warehouse features to algorithm requirements to identify the applicable algorithms for each subproblem.
    
    \item \textbf{Automated pipeline synthesis}: 
    Generation of all valid pipelines from the space of applicable algorithms through combinatory logic synthesis and runtime-efficient execution of all pipelines to identify the best-performing configuration.
\end{enumerate}

Based on these features, CASOP supports a systematic evaluation of algorithmic design choices for a given warehouse context. Rather than relying on manually selected algorithm combinations, the framework identifies applicable components, composes valid pipelines, and evaluates their performance.

We apply CASOP to \nsets{} instance sets covering four problem classes, namely the single picker routing problem with and without scattered storage (SPRP, SPRP-SS), order batching and picker routing problem (OBRP), and the order batching, picker routing, and scheduling problem (OBRSP). To the best of our knowledge, these problems and the corresponding subproblems have not been studied jointly in the literature yet.

The remainder of this paper is organized as follows.
Section~\ref{sec:lit} reviews related work on algorithm selection, semantic modeling, and pipeline generation.
Section~\ref{sec:problem-setting} defines the operational decision problems considered.
Section~\ref{sec:method} presents the CASOP framework.
Section~\ref{sec:experiments} reports on the experimental validation of the framework, followed by a conclusion in Section~\ref{sec:conclusion}.

\section{Related Work}
\label{sec:lit}
In this section, we review work related to ours, namely, algorithm selection approaches in warehouse optimization and related fields, semantic and information modeling for representing systems and algorithms, and automated pipeline synthesis.  
These areas provide the foundation for the context-aware and performance-based selection of algorithms, yet there is currently no approach that achieves an integrated view to facilitate automated warehouse decision-making. 
 
\subsection{Algorithm Selection in Warehouse Order Fulfillment} 
Research on algorithm selection has shown that per-instance learning approaches can significantly outperform single solvers in domains such as SAT \citep{xu2008satzilla} and job-shop scheduling \citep{muller2022schedule}.
By contrast, in warehouse optimization, approaches for algorithm selection remain underdeveloped, although researchers have observed that for problems like picker routing and order batching, the performance of heuristics can depend highly on warehouse characteristics, such as warehouse layout, number of picks or orders, and item distributions.
Early works compared combinations of routing and storage policies \citep{petersen_ii_evaluation_1999}, while later studies expanded these comparisons to include simple constructive batching approaches, zoning, and routing strategies under differing storage strategies, order sizes and batch capacities \citep{van_gils_formulating_2019,bahceci_evaluation_2022}.

\citet{henn_metaheuristics_2010} present a local search approach to the OBRP, which is configured with the routing policies, largest gap, and S-shape. The study confirms previous results on routing algorithm selection, i.e., selecting largest gap when the number of articles to be picked is small and S-shape when there are more articles to be picked.
\citet{van_gils_formulating_2019} evaluate the impact of varying combinations of storage, batching, zoning, and routing strategies. The authors provide an overview of prior work on combinations of different planning problems in warehouse operations.
\citet{luke_single_2024} present a novel, exact solution approach to the single picker routing problem with scattered storage, based on a dynamic programming formulation. Well-known routing heuristics, such as S-shape, return, or largest gap, are formulated with a similar approach. The authors compare different combinations of routing algorithms and storage policies with varying order sizes and  
find that increased scattering of C-articles has a larger effect on tour lengths compared to only scattering A-articles.
 
Beyond these studies, research is moving toward automated or learning-based algorithm selection for warehouse operations problems.  
For example, \citet{cheng_deep_2024} introduce a deep reinforcement learning hyper-heuristic that adaptively selects an order batching strategy from a pool of heuristics. The learned selector significantly outperforms any single heuristic across a variety of test scenarios.
\citet{benavides2025hyper} explore a hyper-heuristic approach to the pod relocation problem in a robotic mobile fulfillment system (RMFS), a form of automated goods-to-person system, effectively treating it as an algorithm selection challenge. 
They develop a portfolio of six heuristics for pod allocation and design 20 different sequence-based hyper-heuristics, which are allocated varying time budgets. The two key contributions are a publicly available simulation framework for RMFS and the implementation of solution strategies. Neither is given for problems in manual warehouse operations. 

Instead, algorithm selection is typically driven by expert knowledge and scenario-specific assumptions, limiting reproducibility and generalization. 
While \citet{hesler_exact_2024} propose a tabular mapping of routing algorithms against instance dimensions such as demand, number of blocks, and item scattering, this representation is limited to a single problem type and is not easily extendable to other decision stages.
Efforts to generalize routing models, for example, by extending the TSP formulation to unconventional layouts such as fishbone or flying-V \citep{wildt_picker_2025}, demonstrate the growing diversity of applicable algorithms, but also highlight the increasing need for a structured applicability model.

\paragraph{\textbf{Gap: No systematic mapping between problem features and solver applicability}}  
Although several studies vary the warehouse characteristics, such as storage policies, order sizes, or item scattering to compare solution approaches \citep{bahceci_evaluation_2022,hesler_exact_2024,luke_single_2024}, no approach explicitly links the warehouse context to algorithm requirements. Current practice remains largely dependent on expert knowledge, which limits reproducibility, transferability, and automation of algorithm selection. 
We address this gap by introducing a mapping procedure based on warehouse features and algorithm requirements.  

\subsection{Information Modeling}
\paragraph{Data Models for Warehouse Operations}
In the context of warehouse optimization, no unified modeling approach to represent warehouse data has been identified.
\citet{oxenstierna_layout-agnostic_2021} were the first to mention the lack of standardized data formats in the context of order batching problems. To alleviate this, they publish an instance benchmark dataset based on the TSPLIB format for TSP. Each instance is represented as a JSON file containing information on the storage locations and obstacles of a digital warehouse model.
Although this dataset is public, the lack of documentation makes it difficult to reproduce or adopt the data model.
In particular, \citet{hesler_exact_2024} made a significant contribution to standardizing instance representations by converting instances from multiple previous works into a unified format. 
They include instances for single picker routing and the order batching and picker routing problem from \cite{bahceci_evaluation_2022,zulj_hybrid_2018,henn_metaheuristics_2010,muter_exact_2015, hesler_exact_2024}. 
Although this serves as a starting point, the instance format of \citet{hesler_exact_2024} encodes several fixed assumptions that constrain both its expressiveness and its adaptability to new problem settings.
For example, a single depot location is assumed, and, besides a generic pick capacity, no information about picker resources is modeled, such as travel or picking speed. Further, the warehouse layout is given in a parameterized fashion, in terms of the number of aisles, number of pick locations, and fixed distances between them. This makes it difficult to describe real-life or unconventional layouts. 

\citet{de_assis_order_2025} introduce a real-world order picking dataset collected from a footwear manufacturer’s warehouse. The dataset includes the warehouse's CAD layout, along with the coordinates of storage locations, aisles, and depots. Further, it contains detailed product and storage information, order lines, orders, and wave assignments. This information is additionally modeled as a UML class diagram. 

\paragraph{Semantic modeling}
Semantic information models for specific domains are essential to communicate information and standardize knowledge. 
A common approach to model semantic information is through ontologies. 
Formal ontologies have been proposed across various domains to standardize knowledge representation and aid complex decision processes. 
In the context of algorithm selection, \citet{kostovska_option_2021} propose the OPTION ontology, a semantically rich data model for benchmarking optimization algorithms.
OPTION provides core objects such as algorithms, problem instances, and performance metrics, enabling annotation and automatic integration of benchmarking data from different platforms.
In intralogistics, \citet{negri_modelling_2017} provide a comprehensive overview of existing ontology frameworks. They extend the Manufacturing Systems Ontology by the most important entities for internal logistics processes. These entities include, for example, transporters, operators, or such that perform storage functions. 
\citet{knoll2019developing} extend this approach by relating the logistical process activity and information flow. 
\citet{klein_ontotlogy_2021} propose an ontology for autonomous warehouses in the context of agile production systems. 

While such ontologies provide a powerful tool to unify existing datasets, the complexity of ontological frameworks may hinder their adoption. For example, OPTION consists of over 300 core classes \citep{kostovska_option_2021}. At the same time, there does not seem to be large adoption, as the published repository shows no activity. 
\citet{knoll2019developing} also highlight that many ontologies in the warehousing area are using different names for the same process and have varying levels of detail, highlighting a lack of standardization. 

\paragraph{Data and Model Cards}
In the context of machine learning, \citet{pushkarna_data_2022} propose a structured way to document datasets in the form of data cards. This is motivated by the increasing complexity of multimodal datasets that are becoming popular and require a standardized representation of their use and intent. 
\citet{mitchell_model_2019} propose model cards to describe meta-information of trained machine learning models. 
For example, a model card for a computer vision model might describe the model’s architecture, the training data it was developed on, its accuracy on various benchmarks, and the contexts for which the model is suited.
\citet{yang_autommlab_2025} demonstrate the flexibility of model cards. They describe trained classification models using model cards.
Based on these cards, models are selected according to the specified task. The model card approach is widely adopted in the machine learning community and is, for example, a key part of one of the most popular machine learning platforms, HuggingFace, where open-source models and data sets are shared. 

\paragraph{\textbf{Gap: Missing semantic and data models for warehouse optimization}}  
While ontological frameworks in logistics \citep{negri_modelling_2017,knoll2019developing} and in algorithm benchmarking \citep{kostovska_option_2021} demonstrate the potential of semantic modeling, they are either too complex for practical adoption or not tailored to warehouse decision problems. There is currently no lightweight semantic or data model that captures both the essential dimensions of warehouse instances (e.g., layout, order characteristics, resources) and the requirements of solution algorithms in a way that supports interoperability and reuse. 
    
\subsection{CLS and Data Pipeline Synthesis}
Software product-line engineering studies how families of related software variants can be derived from shared reusable assets rather than implemented independently \citep{clements2001software,pohl2005software}. 
Within this broader context, the Combinatory Logic Synthesizer (CLS) provides a type-based mechanism for automatically composing valid variants from repositories of typed components \citep{bessai2014combinatory}.
In CLS, each component is assigned a type that encodes both its programmatic interface and semantic properties describing its domain-specific role.
Given a repository of components and a target specification, CLS solves the inhabitation problem by enumerating all valid component compositions that satisfy the specification, thereby generating all feasible solutions (potentially infinitely many) \citep{bessai2014combinatory}.

CLS has been applied across several domains to automate the composition of domain-specific components.
\citet{winkels_simulation_cls} use CLS to synthesize structural variants of discrete-event simulation models for material flow systems. 
Starting from a valid initial model, they represent machines, conveyors, and structural nodes as typed combinators and automatically generate alternative factory layouts that satisfy specified input--output constraints.
\citet{mages2022automatic} employ CLS to generate a product line of manufacturing simulation models from a master model, wrapping machine cells as components and using a graphical user interface to let non-experts configure shop floor layouts; their proof of concept synthesized and simulated 128 configurations in under seven minutes.
\citet{maeckel_scheduling_cls} build a CLS repository of constructive heuristics and dispatching rules for flow shop and job shop problems. 
CLS automatically selects and composes only those algorithms whose intersection types match the given problem class, demonstrating how the framework maps problem characteristics to applicable algorithms.
However, their approach does not consider individual algorithm requirements beyond the machine environment, such as instance-specific feature conditions or data-dependent compatibility constraints.
These applications share a common pattern: domain components are wrapped in semantic types encoding their applicability conditions, and CLS exhaustively enumerates all valid compositions.
This pattern is directly transferable to the composition of decision pipelines for warehouse operations, where algorithms for e.g., order batching, picker routing, and scheduling must be combined.

Selecting well-performing pipelines of algorithm combinations requires building and evaluating many different pipelines.
Given a large design space, this process can quickly become tedious if done manually.
Thus, a systematic automation of pipeline synthesis, evaluation, and selection can offer many efficiency gains.
\citet{meyer_cls-luigi_2024} proposed CLS-Luigi, a framework for automating the generation and execution of analytics pipelines from an algorithm repository.
It combines CLS with Luigi \citep{luigi}, an open-source workflow management system originally developed at Spotify for defining, scheduling, and executing complex dependency-based data processing pipelines.
Unlike existing approaches to automated pipeline synthesis, CLS-Luigi can combine algorithms from different domains, including data preprocessing, machine learning, and decision-making. 
The synthesized pipelines may be non-linear and vary in both length (number of nodes) and structure (flow of data). 
A built-in caching mechanism enables the reuse of intermediate results across pipeline executions. 
Newly added algorithm components can be integrated seamlessly and automatically benefit from the same caching infrastructure.

\paragraph{\textbf{Gap: Lack of automated pipeline synthesis for warehouse optimization}}
While CLS has been successfully applied to synthesize simulation models \citep{winkels_simulation_cls, mages2022automatic}, scheduling heuristics \citep{maeckel_scheduling_cls}, and analytics pipelines \citep{meyer_cls-luigi_2024}, it has not yet been used for warehouse optimization based on a contextual description. 
Current studies still evaluate algorithm combinations manually or restrict to isolated decision problems, whereas the automated generation of valid, instance-specific decision pipelines that span multiple interdependent warehouse operations has not been explored yet.

\section{Decision Problems in Order Fulfillment}
\label{sec:problem-setting}

\begin{figure}[h!]
    \centering
    \includegraphics[width=1\linewidth]{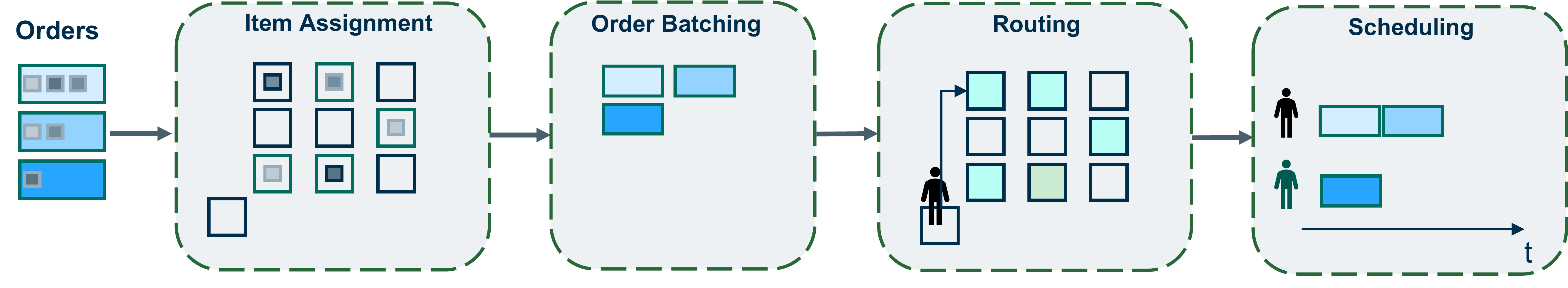}
    \caption{Overview of decomposed decision problems in order fulfillment.}
    \label{fig:warehouse_op}
\end{figure}

Figure~\ref{fig:warehouse_op} illustrates how the subproblems of order fulfillment might be connected. Note that this is only one possible shape; depending on the warehouse context, the problem space may be smaller or larger, and subproblems such as order batching and picker routing could also be solved in an integrated fashion. 
In this section, we present a domain model that structures the data relevant for the order fulfillment subproblems, describe the subproblems in detail, and present an algorithm repository for each subproblem.

\subsection{Domain Model}
\label{sec:domain-model}
The order fulfillment process takes place within a warehouse system composed of storage locations connected by travel aisles.
To encapsulate the data relevant to solving decision problems in order fulfillment, we consider it along five domain objects: \texttt{LayoutDomain}, \texttt{ArticlesDomain}, \texttt{OrdersDomain}, \texttt{ResourcesDomain}, and \texttt{StorageDomain}. 
Each domain object serves as a container for related data models and is annotated by a typed variable $t \in T$ to allow for meta-description of the domain object. Table \ref{tab:notation} summarizes the domain objects and notation used. 
The domain objects and domain models are implemented as Python \texttt{dataclass} objects and are extendable based on the requirements of algorithms.

\paragraph{\texttt{ArticlesDomain}}
Let $\mathcal{A}$ be the set containing all articles present in the warehouse.
Each article $a \in \mathcal{A}$ is characterized by basic physical properties such as its weight $w(a) \in \mathbb{R}_{>0}$ and volume $v(a) \in \mathbb{R}_{>0}$, which define space and capacity requirements for batching.

\paragraph{\texttt{OrdersDomain}}
Let $\mathcal{O}$ denote the set of customer orders.  
Each order $o \in \mathcal{O}$ consists of a set of order lines $\mathcal{L}(o)$.  
Each order line $\ell \in \mathcal{L}(o)$ requests a specific article $a(\ell) \in \mathcal{A}$ in a given quantity $q(\ell) \in \mathbb{N}$.  
Orders may also include temporal attributes such as an arrival time $t^{\mathrm{arr}}(o) \in \mathbb{R}_{\ge 0}$, a due date $t^{\mathrm{due}}(o) \in \mathbb{R}_{\ge 0}$, which are relevant for scheduling decisions, or a completion time $t^{\mathrm{complete}}(o) \in \mathbb{R}_{\ge 0}$ which can be assigned when the order is fulfilled. 
The domain type indicates whether orders are splittable or non-splittable. 

\paragraph{\texttt{ResourcesDomain}}
Fulfillment operations are performed by a set of resources $\mathcal{R}$, such as human pickers or autonomous mobile robots. The domain type categorizes the resource fleet accordingly.   
Each resource $r \in \mathcal{R}$ is specified by its travel speed $s(r) \in \mathbb{R}_{>0}$, and handling time per item $\tau(r) \in \mathbb{R}_{>0}$.  
Resources have a load capacity through an associated pick cart, which is described by $K$ capacity dimensions $\delta = (\delta_1,\ldots,\delta_K)$.
For each resource $r \in \mathcal{R}$ and each capacity dimension
$k \in \{1,\ldots,K\}$, $c_k(r) \in \mathbb{R}_{>0}$ denotes the
capacity of resource $r$ in dimension $k$. The dimension types vary across warehouse settings: for example, \citet{henn_metaheuristics_2010} measure capacity in number of items, while \citet{van_gils_formulating_2019} constrain boxes per cart with each box limited by item count. The pick cart may further specify the 
number of available boxes $n_{\mathrm{box}}(r)$ and whether orders can be mixed 
across boxes.

\paragraph{\texttt{StorageDomain}}
Articles are stored at physical locations $\mathcal{S}$, governed by a storage policy, where each location $s \in \mathcal{S}$ is characterized by coordinates $(x, y)$ in the warehouse layout and may hold a quantity of one or more articles. 
The inventory state at a given point in time is represented by the function $I : \mathcal{A} \times \mathcal{S} \rightarrow \mathbb{Z}_{\ge 0}$, where $I(a, s)$ denotes the inventory level of article $a$ at storage location $s$.  
The total inventory level of article $a$ is given by $\sum_{s \in \mathcal{S}} I(a, s)$.  
The domain type distinguishes between dedicated storage, where each article is assigned to exactly one location, and scattered storage, where articles may be stored at 
multiple locations.

\paragraph{\texttt{LayoutDomain}}
The spatial structure of the warehouse can be described in two complementary 
forms: a parametric description and an explicit spatial representation. 
The parametric description captures the regular structure of conventional parallel-aisle warehouses through the number of aisles $n_a$, blocks $n_b$, pick locations per aisle $n_p$, and distances between them ($d_p$, $d_a$), along with start and end positions $p_s$ and 
$p_e$. 
The explicit spatial representation describes the warehouse as a layout graph $G = (V, E)$, where $V$ denotes pick nodes and depots, $E \subseteq V \times V$ the feasible connections, and $d : V \times V \rightarrow \mathbb{R}_{\ge 0}$ assigns shortest-path distances. 
A predecessor function $\textit{pred}: V \times V \rightarrow V$ supports path reconstruction to obtain the actual pick route.
The domain type distinguishes conventional parallel-aisle layouts from unconventional designs such as fishbone or flying-V \citep{BOYSEN2019396}.

\begin{table}[h!]
\centering
\caption{Summary of notation and domain objects.}
\label{tab:notation}
\small
\begin{tabular}{ll}
\toprule
\textbf{Symbol} & \textbf{Description} \\
\midrule
\multicolumn{2}{l}{\texttt{ArticlesDomain} \textnormal{— type $T \in \{\text{standard}\}$}} \\
\midrule
$\mathcal{A}$ & Set of all articles \\
$a \in \mathcal{A}$ & An article \\
$w(a) \in \mathbb{R}_{>0}$ & Weight of article $a$ \\
$v(a) \in \mathbb{R}_{>0}$ & Volume of article $a$ \\
\midrule
\multicolumn{2}{l}{\texttt{OrdersDomain} \textnormal{— type $T \in \{\text{standard, splittable}\}$}} \\
\midrule
$\mathcal{O}$ & Set of customer orders\\
$o \in \mathcal{O}$ & An order \\
$\mathcal{L}(o)$ & Set of order lines in order $o$\\
$\ell \in \mathcal{L}(o)$ & An order line \\
$a(\ell) \in \mathcal{A}$ & Article requested by order line $\ell$ \\
$q(\ell) \in \mathbb{N}$ & Quantity requested in order line $\ell$ \\
$t^{\mathrm{arr}}(o),\, t^{\mathrm{due}}(o), t^{\mathrm{complete}}(o)\in \mathbb{R}_{\ge 0}$ & Arrival time, due date, and completion time of order $o$ \\
\midrule
\multicolumn{2}{l}{\texttt{ResourcesDomain} \textnormal{— type $T \in \{\text{human, robot, mixed}\}$}} \\
\midrule
$\mathcal{R}$ & Set of resources\\
$r \in \mathcal{R}$ & A resource \\
$s(r) \in \mathbb{R}_{>0}$ & Travel speed of resource $r$ \\
$\tau(r) \in \mathbb{R}_{>0}$ & Handling time per item for resource $r$ \\
$c_k(r) \in \mathbb{R}_{>0}$ & Capacity of resource $r$ in dimension $k$ \\
$K \in \mathbb{N}$ & Number of capacity dimensions \\
$k \in \{1,\ldots,K\}$ & Index of a capacity dimension \\
$\delta = (\delta_1, \ldots, \delta_K)$ & Capacity dimension types (e.g., items, weight, volume, order lines) \\
$n_{\mathrm{box}}(r) \in \mathbb{N}$ & Number of boxes on pick cart \\
\midrule
\multicolumn{2}{l}{\texttt{StorageDomain} \textnormal{— type $T \in \{\text{dedicated, scattered}\}$}} \\
\midrule
$\mathcal{S}$ & Set of storage locations\\
$s \in \mathcal{S}$ & A storage location with coordinates $(x, y)$ \\
$I: \mathcal{A} \times \mathcal{S} \rightarrow \mathbb{Z}_{\ge 0}$ & Inventory state function \\
$I(a, s)$ & Inventory level of article $a$ at location $s$ \\
\midrule
\multicolumn{2}{l}{\texttt{LayoutDomain} \textnormal{— type $T \in \{\text{conventional, unconventional}\}$}} \\
\midrule
$n_a,n_b,n_p \in \mathbb{N}$ & Number of aisles, pick locations, blocks \\
$d_p, d_a \in \mathbb{R}_{>0}$ & Distance between pick locations, aisles \\
$p_s, p_e \in V$ & Start and end (depot) positions \\
$G = (V, E)$ & Warehouse layout graph \\
$V$ & Set of vertices (pick nodes and depots) \\
$E \subseteq V \times V$ & Set of feasible connections \\
$d: V \times V \rightarrow \mathbb{R}_{\ge 0}$ & Distance function\\
$\textit{pred}: V \times V \rightarrow V$ & Predecessor function for path reconstruction \\
\bottomrule
\end{tabular}
\end{table}

\subsection{Subproblems of Order Fulfillment}
\label{sec:subproblems}
We decompose order fulfillment into four subproblems: item assignment, order batching, picker routing, and picker scheduling.
We describe each of these subproblems in detail below.

\textbf{Item Assignment (IA).} In warehouses with scattered storage policies, where articles may be stored at multiple locations simultaneously, the item assignment problem determines from which specific storage location(s) each requested article should be picked. This stage resolves each order line $\ell \in \mathcal{L}(o)$ to one or more storage locations $s \in \mathcal{S}$ based on the inventory function $I(a(\ell), s)$, such that the requested quantity $q(\ell)$ can be fulfilled. When a single location has insufficient inventory, the quantity may be split across multiple locations.
The output is a set of resolved pick positions $\mathcal{P}$, where each $p \in \mathcal{P}$ specifies a pick node, the quantity to retrieve, and its associated order line. 

\textbf{Order Batching (B).} To reduce travel time, multiple customer orders can be combined into batches that are retrieved in a single picker tour. The batching problem groups orders $o \in \mathcal{O}$ into batches $b \in \mathcal{B}$, subject to capacity constraints imposed by the picker cart capacities $c_k(r)$, $k \in \{1,\ldots,K\}$. Each batch forms a pick list containing all order lines from its constituent orders. 
Every order $o$ may have an arrival time $t^{\mathrm{arr}}(o)$ that indicates when the order becomes known in the system, a due date $t^{\mathrm{due}}(o)$ that specifies the time at which the picking process of the order has to be finished, and a completion time $t^{\mathrm{complete}}(o)$ that captures the time at which the picking process has actually finished. 

\textbf{Picker Routing (R).} For each pick list, the routing problem determines the sequence in which pick locations are visited and the path through the warehouse. This is often modeled as a variant of the TSP with the objective of finding a tour $\sigma$ that visits all required pick locations while minimizing total travel distance. The output specifies both the route through the warehouse layout graph $G = (V, E)$ and the sequence in which items are picked.

\textbf{Scheduling (S).} 
Picker scheduling determines when each pick list or tour is executed and which resource handles it.
In a multi-picker system, scheduling comprises both sequencing and resource assignment.
Sequencing determines the sequence in which the tours are executed, while assignment allocates tours to available resources. 
In a single-picker system, scheduling reduces to sequencing.
The output is a schedule consisting of an assignment function $\rho: \mathcal{B} \rightarrow \mathcal{R}$, which maps each batch $b \in \mathcal{B}$ to a resource $r \in \mathcal{R}$, and a sequence $\pi$, which specifies the execution sequence of batches together with their start and completion times.

\subsection{Optimization Objectives}
The choice of the overall optimization objective in the order fulfillment context depends on operational priorities and available data. Common objectives in order fulfillment include:

\begin{itemize}
    \item Distance: Total travel distance across all pick tours, $\sum_{b \in \mathcal{B}} d(\sigma_b)$, where $\sigma_b$ denotes the tour for batch $b$ and $d(\sigma_b)$ denotes the total travel distance of this tour.
    
    \item Makespan: Time from the start of the first pick tour until completion of the last, $\max_{r \in \mathcal{R}} t^{\mathrm{end}}_r$, where $t^{\mathrm{end}}_r$ denotes the completion time of resource $r$, accounting for travel speed $s(r)$ and handling time $\tau(r)$.
    
    \item Tardiness: Sum of delays beyond due dates, $\sum_{o \in \mathcal{O}} \max(0, t^{\mathrm{complete}}(o) - t^{\mathrm{due}}(o))$, where orders completed before their due date contribute zero.
    
    \item On-Time rate: Percentage of orders completed by their due date, $\frac{|\{o \in \mathcal{O} \mid t^{\mathrm{complete}}(o) \leq t^{\mathrm{due}}(o)\}|}{|\mathcal{O}|} \times 100\%$.
\end{itemize}

\subsection{Algorithms for Order Fulfillment}
\label{sec:algos} 

To the best of our knowledge, no open-source implementations of the algorithms reviewed in this paper exist. We therefore provide a dedicated repository of reusable, modular implementations covering the decision problems defined in Section~\ref{sec:problem-setting}, available at 
\url{https://github.com/kit-dsm/ware_ops_algos}.

The repository currently contains 22 algorithm implementations: five item assignment approaches, seven batching approaches, eight routing approaches, one integrated batching-and-routing approach, and one scheduling approach. 
All algorithms implement a shared \texttt{Algorithm} base class parameterized by input and output types. 
Each algorithm operates on the domain objects introduced in Section~\ref{sec:domain-model} and returns a typed \texttt{SolutionObject} that downstream algorithms can consume. 
The repository is organized into dedicated modules for each decision problem, following a consistent pattern: an abstract base class defines the interface, and concrete implementations specialize it. 
In Tables~\ref{tab:ia-algos}--\ref{tab:algo-sched}, indented entries denote concrete implementations of the base class listed above.  We use 'H', 'MH', and 'E' to indicate whether an algorithm is heuristic, metaheuristic, or exact. Algorithms marked with \algstar{} are used in the framework validation described in Section~\ref{sec:experiments}.

\subsubsection{Item Assignment}
The \texttt{item\_assignment} module contains implementations of five item assignment approaches, summarized in Table~\ref{tab:ia-algos}. 
The module covers both dedicated and scattered storage settings.
For dedicated storage, item assignment is fixed because each article is associated with a single storage position.
As a simple baseline solution for such cases, we implemented GIA, which greedily selects the storage location with the highest inventory level first.
For scattered storage, several storage positions may be feasible for the same requested article, and an assignment rule must select one or more of them.
For the scattered case, we implement the heuristic assignment rules MinMaxIA, MinMinIA, and SPIA based on the priority-based approach presented in \citet{weidinger2018picker} and NNIA, which is adopted from \citet{weidinger_picker_2019}.  
GIA and NNIA are standalone constructive heuristics. 
MinMaxIA, MinMinIA, and SPIA are variants of \texttt{PriorityItemAssignment}, a shared base class for penalty-based priority rules. 
SPIA is parameterized by a routing algorithm that serves as a distance evaluator.

This yields twelve executable item-assignment configurations: four standalone configurations (GIA, NNIA, MinMaxIA, and MinMinIA) and eight SPIA configurations, one for each available routing algorithm.
Each algorithm returns an \texttt{ItemAssignmentSolution} containing $\mathcal{P}$.


\begin{algorithmtable}
{Item assignment algorithms.}
{tab:ia-algos}
\toprule
\textbf{Algorithm} & \textbf{Type} & \textbf{Description} \\
\midrule
\texttt{GreedyItemAssignment} (GIA) 
& H 
& Assigns items from locations with the highest inventory level first until the requested quantity is satisfied. \\

\texttt{NearestNeighborItemAssignment} (NNIA)\algstar{}  
& H 
& Iteratively selects the nearest location until all demand is fulfilled \citep{weidinger_picker_2019}. \\
\midrule
\texttt{PriorityItemAssignment} 
& H 
& Base class for item assignment based on penalty-based priority rules \citep{weidinger2018picker}. \\

\algindent\texttt{MinMaxItemAssignment} (MinMaxIA)\algstar{} 
& 
& Computes penalties as the maximum distance to already-selected locations. \\

\algindent\texttt{MinMinItemAssignment} (MinMinIA)\algstar{} 
& 
& Computes penalties as the minimum distance to already-selected locations. \\

\algindent\texttt{SinglePosItemAssignment} (SPIA)\algstar{} 
& 
& Uses the distance from a fixed reference position. \\
\bottomrule
\end{algorithmtable}

\subsubsection{Batching}
The \texttt{batching} module contains four batching implementations, summarized in Table~\ref{tab:algo-batch}. 
We include constructive, savings-based, seed-based, and local-search approaches. 
Constructive and metaheuristic approaches are discussed by~\citet{henn_metaheuristics_2010}, \citet{Koch2014}, and \citet{pinto_classification_2023}.
Learning-based batching methods have recently been explored by~\citet{beeks2022deep} and~\citet{cals2021solving}.

\texttt{PriorityBatching} is a greedy constructive method parameterized by a sorting criterion. We implement four sorting rules: OrderNrFiFo, FiFo, DueDate, and Random.
\texttt{SavingsBatching} requires a routing evaluator for computing savings and can be configured with any of the eight routing classes described in the next subsection. This generates eight different configurations of the \texttt{SavingsBatching} algorithm.
\texttt{SeedBatching} is parameterized by three \texttt{SeedCriteria} and two \texttt{SimilarityMeasure} options as specified in Table~\ref{tab:seed-config}, resulting in six configurations.
\texttt{LocalSearchBatching} requires both a constructive batching method for initialization and a routing evaluator. Combined with four constructive batching approaches (FiFo, OrdNrFiFo, DueDate, RAND) and eight routing classes, this yields 32 configurations.

These implementations yield 50 executable batching configurations: four priority-based configurations, eight savings-based configurations, six seed-based configurations, and 32 local-search configurations.
Each algorithm returns a \texttt{BatchingSolution} containing $\mathcal{B}$ and the derived pick lists.


\begin{algorithmtable}[H]
{Batching algorithms.}
{tab:algo-batch}
\toprule
\textbf{Algorithm} & \textbf{Type} & \textbf{Description} \\
\midrule
\texttt{PriorityBatching} 
& H 
& Greedy constructive batching by sorting criterion. \\

\quad \texttt{OrderNrFiFo}\algstar{} 
& 
& By ascending order number. \\

\quad \texttt{FiFo}\algstar{} 
& 
& By arrival date $t^{\mathrm{arr}}(o)$. \\

\quad \texttt{DueDate}\algstar{} 
& 
& By due date $t^{\mathrm{due}}(o)$. \\

\quad \texttt{Random}\algstar{} 
& 
& Random order with fixed seed. \\
\midrule
\texttt{SavingsBatching}\algstar{} 
& H 
& Merges batch pairs with largest routing distance saving~\citep{clarke_and_wright}. Configurable routing evaluator (Table~\ref{tab:algo-route}). \\
\midrule
\texttt{SeedBatching}\algstar{} 
& H 
& Builds batches around seed orders. Configurable seed selection and similarity measure (Table~\ref{tab:seed-config}). \\
\midrule
\texttt{LocalSearchBatching}\algstar{} 
& MH 
& Iterative improvement via SWAP/SHIFT operators~\citep{henn_metaheuristics_2010}. Configurable constructive initialization; see \texttt{PriorityBatching}, and routing evaluator (Table~\ref{tab:algo-route}). \\
\bottomrule
\end{algorithmtable}


\begin{configtable}[H]
{Configuration options for \texttt{SeedBatching}.}
{tab:seed-config}
\toprule
\textbf{Category} & \textbf{Option} & \textbf{Description} \\
\midrule
Seed selection & \texttt{RANDOM} & Random seed order. \\
& \texttt{MOST\_POSITIONS} & Order with the most positions. \\
& \texttt{FEWEST\_POSITIONS} & Order with the fewest positions. \\
& \texttt{CLOSEST\_TO\_DEPOT} & Order whose nearest pick node is closest to $p_s$. \\
\midrule
Similarity measure & \texttt{SHARED\_ARTICLES} & Orders with more shared articles are considered more similar. \\
& \texttt{MIN\_DISTANCE} & Minimum pairwise pick-node distance between orders. \\
\bottomrule
\end{configtable}

\subsubsection{Routing}
The \texttt{routing} module contains eight standalone routing implementations, summarized in Table~\ref{tab:algo-route}. The layout-dependent heuristics follow established routing rules for parallel-aisle warehouses, as evaluated by~\citet{petersen_evaluation_1997}. 
We also implement nearest-neighbor routing, two exact TSP-based formulations, and the dynamic programming
approach of~\citet{ratliff_order-picking_1983} as extended by~\citet{hesler_exact_2024} for single-block parallel-aisle layouts with scattered storage. 
For a broader overview of TSP-based picker-routing models, we refer to~\citet{bock2025survey}. 

In addition, we provide CBR, an integrated batching-and-routing approach. 
Joint order batching and picker routing has been studied as an integrated optimization problem by~\citet{valle_optimally_2017}; a more general column-generation-based approach is presented by~\citet{briant_efficient_2020}.
Our implementation operates directly on $\mathcal{O}$ and jointly determines batching and routing via a capacity-constrained TSP formulation.

Each routing algorithm returns a \texttt{RoutingSolution} containing $\sigma$ and $d(\sigma)$. This allows for the reconstruction and visualization of the picker tours.  
\begin{algorithmtable}
{Routing algorithms.}
{tab:algo-route}
\toprule
\textbf{Algorithm} & \textbf{Type} & \textbf{Description} \\
\midrule
\texttt{HeuristicRouting} 
& H 
& Base class for routing heuristics. \\

\quad \texttt{SShapeRouting} (SShape)\algstar{} 
& 
& Traverse aisles in an S-pattern. \\

\quad \texttt{ReturnRouting} (RET)\algstar{} 
& 
& Enter each aisle and return after the last pick. \\

\quad \texttt{MidpointRouting} (MP)\algstar{} 
& 
& Split the warehouse at a horizontal midpoint, processing the lower and upper parts separately. \\

\quad \texttt{LargestGapRouting} (LG)\algstar{} 
& 
& Enter each aisle only up to the largest empty gap. \\

\quad \texttt{NearestNeighborRouting} (NN)\algstar{} 
& 
& Greedily visit the nearest unvisited pick location. \\
\midrule
\texttt{ExactRouting} 
& E 
& Base class for exact routing models. \\

\quad \texttt{ExactTSPRoutingDistance} (TSP-D) 
& 
& TSP with subtour elimination. \\

\quad \texttt{ExactTSPRoutingTime} (TSP-T) 
& 
& TSP including travel speed and handling time. \\
\midrule
\texttt{RatliffRosenthalRouting} (RR)\algstar{} 
& E 
& DP for single-block rectangular warehouses. \\
\midrule
\texttt{RoutingBatching} 
& E 
& Base class for integrated batching-and-routing approaches. \\

\quad \texttt{CombinedBatchingRouting} (CBR)\algstar{} 
& 
& Capacity-constrained TSP for integrated batching-and-routing. \\
\bottomrule
\end{algorithmtable}

\subsubsection{Scheduling} 
Batches or pick tours are usually sequenced using dispatching rules from the scheduling literature. 
We implement list-scheduling, an approach introduced by~\citet{graham_1969} for multiprocessor scheduling, with four dispatching rules summarized in Table~\ref{tab:algo-sched}. 

The implementation operates on routing solutions $\sigma$, orders $\mathcal{O}$ with temporal attributes $t^{\mathrm{arr}}(o)$ and $t^{\mathrm{due}}(o)$, and resources $\mathcal{R}$ with travel speed $s(r)$ and handling time $\tau(r)$. 
It maintains a priority-ordered list of tours and repeatedly assigns the highest-priority tour to the earliest available picker. The dispatching rule defines the priority.

This yields four executable scheduling configurations: shortest processing time (SPT), longest processing time (LPT), earliest due date (EDD), and earliest release date (ERD).
Each algorithm returns a \texttt{SchedulingSolution} containing the assignment 
$\rho: \mathcal{B} \rightarrow \mathcal{R}$, the execution sequence $\pi$, and start and completion times for each tour.

\begin{algorithmtable}
{Scheduling algorithms.}
{tab:algo-sched}
\toprule
\textbf{Algorithm} & \textbf{Type} & \textbf{Description} \\
\midrule
\texttt{ListScheduling} 
& H 
& Base class for priority-based list scheduling. \\

\quad \texttt{SPTScheduling} (SPT)\algstar{} 
& 
& Shortest processing time first. \\

\quad \texttt{LPTScheduling} (LPT)\algstar{} 
& 
& Longest processing time first. \\

\quad \texttt{EDDScheduling} (EDD)\algstar{} 
& 
& Earliest due date first. \\

\quad \texttt{ERDScheduling} (ERD) 
& 
& Earliest release date first. \\
\bottomrule
\end{algorithmtable}

\subsection{Configuration Space of the Algorithm Repository}
The algorithm repository described in Section~\ref{sec:algos} contains both fixed and parameterized implementations.
We refer to each fully specified variant as an executable algorithm configuration, and write $\mathit{IA}$, $\mathit{B}$, $\mathit{R}$, $\mathit{BR}$, and $\mathit{S}$ for the sets of item-assignment, batching, standalone routing, integrated batching-and-routing, and scheduling configurations.
Across all subproblems, the repository contains 75 executable algorithm configurations: $|\mathit{IA}| = 12$, $|\mathit{B}| = 50$, $|\mathit{R}| = 8$, $|\mathit{BR}| = 1$, and $|\mathit{S}| = 4$.
This count describes the available algorithm configurations prior warehouse-specific filtering and composition into complete optimization pipelines.
A purely sequential order batching and routing setting with item assignment already admits $|\mathit{IA}| \cdot |\mathit{B}| \cdot |\mathit{R}| = 12 \cdot 50 \cdot 8 = 4{,}800$ possible combinations.
If scheduling is included, the corresponding sequential space grows to $|\mathit{IA}| \cdot |\mathit{B}| \cdot |\mathit{R}| \cdot |\mathit{S}| = 12 \cdot 50 \cdot 8 \cdot 4 = 19{,}200$ combinations.

However, valid pipeline construction is not a cartesian product over all stages.
Not every problem class requires every subproblem.
A pure routing problem does not require batching or scheduling, while an integrated batching-and-routing configuration from $\mathit{BR}$ bypasses the separate batching and routing stages, adding $|\mathit{IA}| \cdot |\mathit{BR}|$ combinations rather than entering the product above.
The applicable algorithms also depend on the warehouse context and the available data.
For example, item-assignment variants are only relevant when the storage model allows alternative pick locations, scheduling requires temporal order information and multiple resources, and some routing algorithms require a specific layout representation or layout class.

Thus, the relevant configuration space is instance-dependent.
Determining which algorithm configurations are applicable and composing them into valid executable pipelines is the key challenges addressed by the CASOP framework introduced in the following section.

\section{\frameworkname}
\label{sec:method}
We present the framework coined \emph{\frameworklong{}} (\frameworkname) that enables the automated mapping of warehouse characteristics to compatible algorithmic approaches, the synthesis of valid algorithm pipelines, and their systematic evaluation.

\subsection{CASOP Overview}
The  CASOP framework consists of the following five core building blocks, which are illustrated in Figure~\ref{fig:3d4l-framework} and discussed in detail in the following subsections:
\begin{figure}[h!]
    \centering
    \includegraphics[width=1\linewidth]{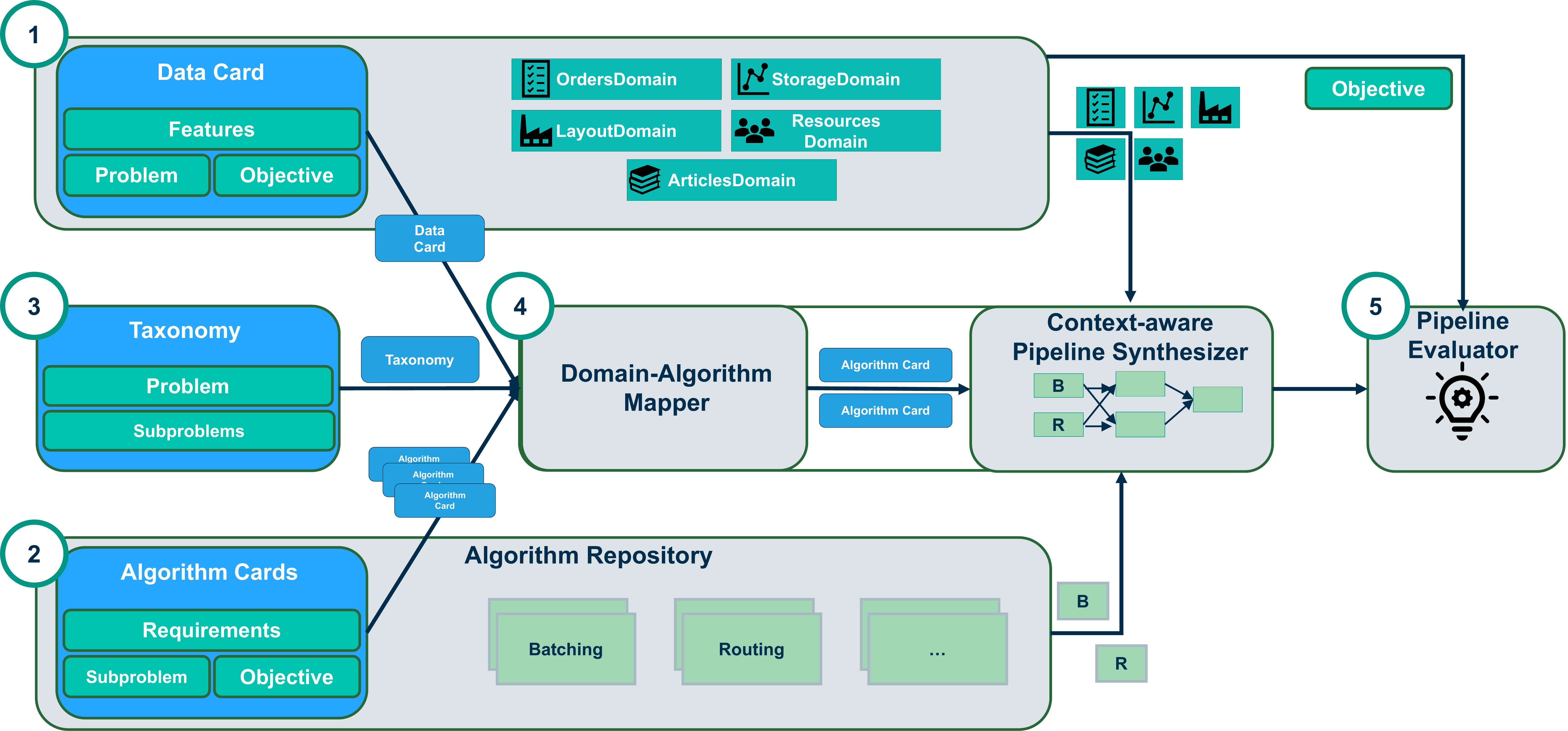}
    \caption{Core building blocks of the \frameworkname framework. Warehouse data is semantically structured into domain models described in the form of a data card (1). Requirements of algorithms from a repository are described in the form of algorithm cards (2). A taxonomy describes the problem space and which subproblems a problem consists of (3). Algorithm card, data card, and taxonomy are used by the domain-algorithm mapping component to obtain applicable algorithms, which are then composed into context-aware pipelines (4). Finally, all resulting pipelines are evaluated to select the best-performing configuration (5).}
    \label{fig:3d4l-framework}
\end{figure}

\begin{itemize}
\item[(1)]\textbf{Data Cards and Domain Objects.}
Warehouse information is organized into domain objects, such as layout, articles, orders, resources, and storage.
Each domain object contains contextual information, such as the layout type, and data features, such as the number of aisles or picker capacity.
A data card provides a semantic description of these domain objects and declares the available features, the problem class to be solved, and the optimization objective.

\item[(2)]\textbf{Algorithm Cards and Algorithm Repository.}
The algorithm repository contains implementations of item assignment, batching, routing, and scheduling algorithms, i.e., the algorithms presented in Section~\ref{sec:algos}.
Each algorithm is described by an algorithm card that specifies the subproblem it addresses, its objective, and its requirements with respect to the warehouse data and context.

\item[(3)]\textbf{Taxonomy.}
The taxonomy defines the problem space considered by the framework.
It maps each problem class to the corresponding subproblems.
Depending on the warehouse setting and problem class, only a subset of subproblems may be relevant, and some subproblems may be solved sequentially or in an integrated fashion.

\item[(4)]\textbf{Pipeline Synthesizer.}
The pipeline synthesizer generates executable optimization pipelines subject to the context provided by the data card, algorithm cards, and taxonomy.
It comprises two parts.
First, the domain-algorithm mapper identifies all algorithms applicable to the given warehouse setting. 
It filters algorithms by problem type using the taxonomy and checks whether the required domain types, features, and feature constraints specified in the algorithm cards are satisfied by the context provided through the data card.
Second, the applicable algorithms are synthesized into context-aware pipelines, represented as directed acyclic graphs in which nodes correspond to algorithms or data sources and edges capture data dependencies between decision stages.

\item[(5)]\textbf{Pipeline Evaluator.}
The pipeline evaluator evaluates the executed pipelines and assesses them according to the objective specified in the data card.
The result is a ranked set of valid pipelines and the best-performing configuration for the given warehouse context.
\end{itemize}

\subsection{Data Cards and Domain Objects}
\label{sec_domain_model}
A data card $\mathcal{I}$ captures the semantic description of a 
warehouse setting. It is defined by:
\[
  \mathcal{I} \;=\; \langle P,\, Z,\, \mathcal{D}_1, \ldots, 
  \mathcal{D}_5 \rangle
\]
where $P$ is the problem class, $Z$ is the objective, and 
$\mathcal{D}_1, \ldots, \mathcal{D}_5$ correspond to the five data 
domains introduced in Section~\ref{sec:domain-model}. 
Each domain object $\mathcal{D}_i = \langle T_i, F_i \rangle$ expresses the categorical type $T_i$ (e.g., conventional, scattered, human), and features $F_i$ of the related domain models, as defined in Table~\ref{tab:notation}.

Data cards are structured hierarchically in YAML, mirroring the domain composition. Features are listed by name. 
When a feature carries a concrete value (e.g., \texttt{n\_blocks: 2}), this value can be used for constraint evaluation in algorithm cards.
Features listed without a value signal availability only; their presence indicates that the corresponding data exists in the warehouse setting.
Figure~\ref{fig:card-examples}(a)  shows an 
example for the \texttt{Kris} instance set \citep{valle_optimally_2017}.

\begin{figure*}[t]
\centering
\begin{minipage}[t]{0.47\textwidth}
\centering
\textbf{(a) Data card}
\vspace{0.3em}
\begin{lstlisting}[style=cardstyle]
name: example_data_card
problem_class: OBRSP
objective: distance
LayoutDomain:
  type: conventional
  features:
    - start_node
    - end_node
    - n_blocks: 2
    - graph
    - distance_matrix
ArticlesDomain:
  type: standard
  features:
    - article_id
    - weight
    - volume
OrdersDomain:
  type: standard
  features:
    - order_number
    - order_date
    - amount
    - article_id
ResourcesDomain:
  type: human
  features:
    - speed
    - time_per_pick
    - capacities
    - dimensions_type: [items]
StorageDomain:
  type: dedicated
  features:
    - amount
    - article_id
\end{lstlisting}
\end{minipage}
\hfill
\begin{minipage}[t]{0.47\textwidth}
\centering
\textbf{(b) Algorithm card}
\vspace{0.3em}
\begin{lstlisting}[style=cardstyle]
model_name: RatliffRosenthal
problem_type: routing
objective: distance
requirements:
  LayoutDomain:
    type:
      - conventional
    features:
      - start_node
      - end_node
      - closest_node_to_start
      - min_aisle_position
      - max_aisle_position
      - n_pick_locations
      - n_aisles
      - dist_pick_locations
      - dist_bottom_to_pick_location
      - n_blocks
    constraints:
      n_blocks:
        equals: 1
  ResourcesDomain:
    type:
      - any
  OrdersDomain:
    type:
      - standard
  StorageDomain:
    type:
      - any
  ArticlesDomain:
    type:
      - any
\end{lstlisting}
\end{minipage}
\caption{Examples of a data card and an algorithm card. The data card describes an instance from the \textit{Kris} dataset, whereas the algorithm card describes the requirements of the \texttt{RatliffRosenthalRouting} algorithm.}
\label{fig:card-examples}
\end{figure*}

\subsection{Algorithm Cards}
\label{sec:algo-cards}
 
Each algorithm in the repository introduced in Section~\ref{sec:algos} is accompanied by an algorithm card. An algorithm card $\alpha \in \mathbb{A}$ is defined by:
\[
  \alpha \;=\; \langle R_{\alpha},\, P_{\alpha},\, Z_{\alpha} \rangle
\]
where $P_{\alpha}$ is the subproblem addressed, $Z_\alpha$ is the optimization objective (empty for constructive methods without an explicit objective), and $R_{\alpha} = \langle T_{\alpha}, 
F_{\alpha}, C_{\alpha} \rangle$ specifies the algorithm requirements: the required domain types 
($T_{\alpha}$), features ($F_{\alpha}$), and constraints ($C_{\alpha}$). 
Constraints can include comparison and set relations over features such as \texttt{equals}, \texttt{greater\_than}, or \texttt{in}, as well as logical combinations using \texttt{and} and \texttt{or}. 
This allows modeling requirements that go beyond feature presence.

Algorithm requirements, subproblem, and objective are specified as YAML-formatted algorithm cards. 
Figure~\ref{fig:card-examples}(b) shows an example for the \texttt{RatliffRosenthalRouting} algorithm \citep{ratliff_order-picking_1983,hesler_exact_2024}.

\subsection{Taxonomy}
\label{sec:taxonomy}
\begin{figure}[htbp]
    \centering
    \resizebox{0.82\linewidth}{!}{%
    \begin{tikzpicture}[
        taxnode/.style={
            draw=casopDark,
            fill=casopLightBlue,
            line width=0.55pt,
            rounded corners=0.6pt,
            minimum width=12mm,
            minimum height=4mm,
            inner sep=0pt,
            align=center,
            font=\sffamily\bfseries\scriptsize
        },
        subnode/.style={
            taxnode,
            fill=casopBlue,
            minimum width=10mm,
            minimum height=3.7mm
        },
        taxline/.style={
            draw=casopDark,
            line width=0.55pt,
            line cap=round,
            line join=round
        }
    ]

    \node[taxnode] (P) at (0,0) {P};

    \node[taxnode] (PRP)   at (-4.8,-0.85) {PRP};
    \node[taxnode] (OBP)   at (-1.6,-0.85) {OBP};
    \node[taxnode] (OBRP)  at ( 1.6,-0.85) {OBRP};
    \node[taxnode] (OBRSP) at ( 4.8,-0.85) {OBRSP};

    \coordinate (root)    at ( 0.0,-0.28);
    \coordinate (busleft) at (-4.8,-0.28);
    \coordinate (busright)at ( 4.8,-0.28);

    \coordinate (cPRP)    at (-4.8,-0.28);
    \coordinate (cOBP)    at (-1.6,-0.28);
    \coordinate (cOBRP)   at ( 1.6,-0.28);
    \coordinate (cOBRSP)  at ( 4.8,-0.28);

    \draw[taxline] (P.south) -- (root);
    \draw[taxline] (busleft) -- (busright);

    \draw[taxline] (cPRP)   -- (PRP.north);
    \draw[taxline] (cOBP)   -- (OBP.north);
    \draw[taxline] (cOBRP)  -- (OBRP.north);
    \draw[taxline] (cOBRSP) -- (OBRSP.north);

    \node[subnode] (prpIA) at (-4.8,-1.55) {IA};
    \node[subnode] (prpR)  at (-4.8,-2.18) {R};

    \draw[taxline] (PRP.south) -- (prpIA.north);
    \draw[taxline] (prpIA.south) -- (prpR.north);

    \node[subnode] (obpIA) at (-1.6,-1.55) {IA};
    \node[subnode] (obpB)  at (-1.6,-2.18) {B};

    \draw[taxline] (OBP.south) -- (obpIA.north);
    \draw[taxline] (obpIA.south) -- (obpB.north);

    \node[subnode] (obrpIA) at (1.6,-1.55) {IA};
    \node[subnode] (obrpB)  at (1.0,-2.18) {B};
    \node[subnode] (obrpR)  at (1.0,-2.81) {R};
    \node[subnode] (obrpBR) at (2.2,-2.18) {BR};

    \draw[taxline] (OBRP.south) -- (obrpIA.north);

    \coordinate (forkOBRP) at (1.6,-1.80);
    \draw[taxline] (obrpIA.south) -- (forkOBRP);
    \draw[taxline] (forkOBRP) -| (obrpB.north);
    \draw[taxline] (forkOBRP) -| (obrpBR.north);

    \draw[taxline] (obrpB.south) -- (obrpR.north);

    \node[subnode] (obrspIA)  at (4.8,-1.55) {IA};
    \node[subnode] (obrspB)   at (4.2,-2.18) {B};
    \node[subnode] (obrspR)   at (4.2,-2.81) {R};
    \node[subnode] (obrspS)   at (4.2,-3.44) {S};

    \node[subnode] (obrspBR)  at (5.4,-2.18) {BR};
    \node[subnode] (obrspSBR) at (5.4,-2.81) {S};

    \draw[taxline] (OBRSP.south) -- (obrspIA.north);

    \coordinate (forkOBRSP) at (4.8,-1.80);
    \draw[taxline] (obrspIA.south) -- (forkOBRSP);
    \draw[taxline] (forkOBRSP) -| (obrspB.north);
    \draw[taxline] (forkOBRSP) -| (obrspBR.north);

    \draw[taxline] (obrspB.south) -- (obrspR.north);
    \draw[taxline] (obrspR.south) -- (obrspS.north);

    \draw[taxline] (obrspBR.south) -- (obrspSBR.north);

    \end{tikzpicture}%
    }
    \caption{Taxonomy implemented in CASOP. Problem (P); item assignment (IA); batching (B); routing (R); scheduling (S); picker routing problem (PRP); order batching problem (OBP); order batching and routing problem (OBRP); order batching, routing, and scheduling problem (OBRSP).}
    \label{fig:taxonomy}
\end{figure}
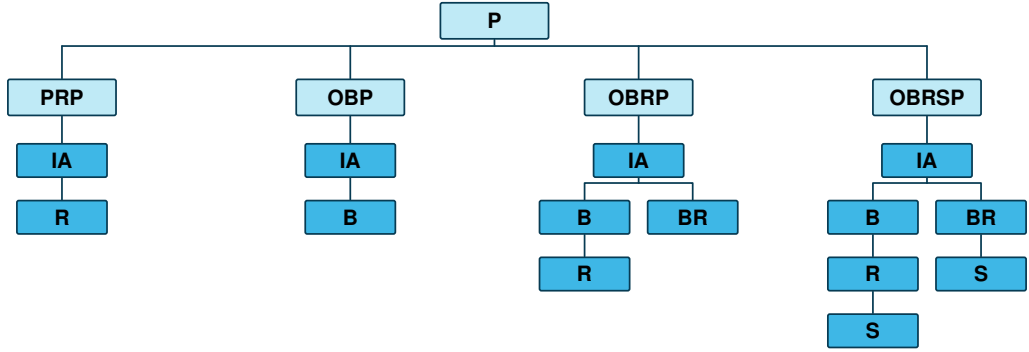

The problem taxonomy $T_P$ formalizes the mapping from each problem class $P$ to its constituent subproblems.
The taxonomy is given in Figure~\ref{fig:taxonomy}. It is based on the classification of order batching problems recently proposed by \citet{pardo_order_2024}. To allow a broader range of problems to be covered, we extend it by the picker routing problem (PRP) and the IA subproblem, which are not covered in their work. 

In our taxonomy, the PRP requires only item assignment and routing; the order batching problem (OBP) adds batching, and the OBRP combines item assignment, order batching, and picker routing, with the option to solve batching and routing jointly through an integrated formulation (BR). The OBRSP additionally includes scheduling.

\subsection{Pipeline Synthesizer}
The pipeline synthesizer generates executable optimization pipelines for a given warehouse setting. It uses the data card $\mathcal{I}$, the algorithm cards $\mathbb{A}$, and the taxonomy $T_P$ to first identify applicable algorithms, then compose them into valid context-aware pipelines, and execute them.

\subsubsection{Domain-Algorithm Mapping}
\label{sec:mapping}

\frameworkname includes a mapping mechanism that determines which algorithms from the repository $\mathbb{A}$ are applicable to a given data card $\mathcal{I}$. 
The mapping produces the subset $\mathbb{A}^* \subseteq \mathbb{A}$ of applicable algorithms. 
The procedure is implemented by the \texttt{DomainAlgoMapper} class, which takes $\mathbb{A}$, $\mathcal{I}$, and $T_P$ as input and returns $\mathbb{A}^*$.
It proceeds in two stages.

\begin{itemize}

\item[(1)]\textbf{Problem-type filtering.}
An algorithm described by algorithm card $\alpha = \langle R_\alpha, P_\alpha, Z_\alpha \rangle$ passes the first stage if its declared subproblem $P_\alpha$ corresponds to the problem class $P$ declared in $\mathcal{I}$, or 
to one of its subproblems as defined by the taxonomy $T_P$ (see Section~\ref{sec:taxonomy}). 
For example, if $\mathcal{I}$ declares $P = \text{OBRP}$, then algorithms for the subproblems IA, B, R, and BR are admitted, while scheduling algorithms are filtered out.

\item[(2)]\textbf{Requirement matching.}
Each remaining algorithm is checked against the domain objects $\mathcal{D}_1, \ldots, \mathcal{D}_5$ of $\mathcal{I}$. 
Specifically, for each domain $\mathcal{D}_i = \langle T_i, F_i \rangle$ referenced by the algorithm requirements $R_\alpha = \langle T_\alpha, F_\alpha, C_\alpha \rangle$, the 
framework verifies:
\begin{enumerate}
  \item[(a)] Type compatibility: The required domain type matches the data card's type, i.e., $T_i \in T_\alpha$, where \texttt{any} indicates no restriction.
  \item[(b)] Feature availability: All required features are present, i.e., $F_\alpha \subseteq F_i$.
  \item[(c)] Constraint satisfaction: All feature constraints $C_\alpha$ are satisfied by the concrete feature values in $\mathcal{I}$.
\end{enumerate}
An algorithm $\alpha$ is applicable if and only if conditions (a)-(c) hold for every referenced domain.

\end{itemize}

\subsubsection{Context-Aware Pipelines}
\label{sec:pipeline-generation}

The synthesizer builds valid algorithm pipelines for the active problem class, executes them, and ranks their results. 
It uses CLS-Luigi, which combines CLS with Luigi. 
CLS takes a repository of typed components and a target component, and searches for component combinations that can produce this target.
Luigi executes the resulting combinations as directed acyclic task graphs and caches intermediate results.

\begin{figure}[h!]
\centering
\includegraphics[width=1\linewidth]{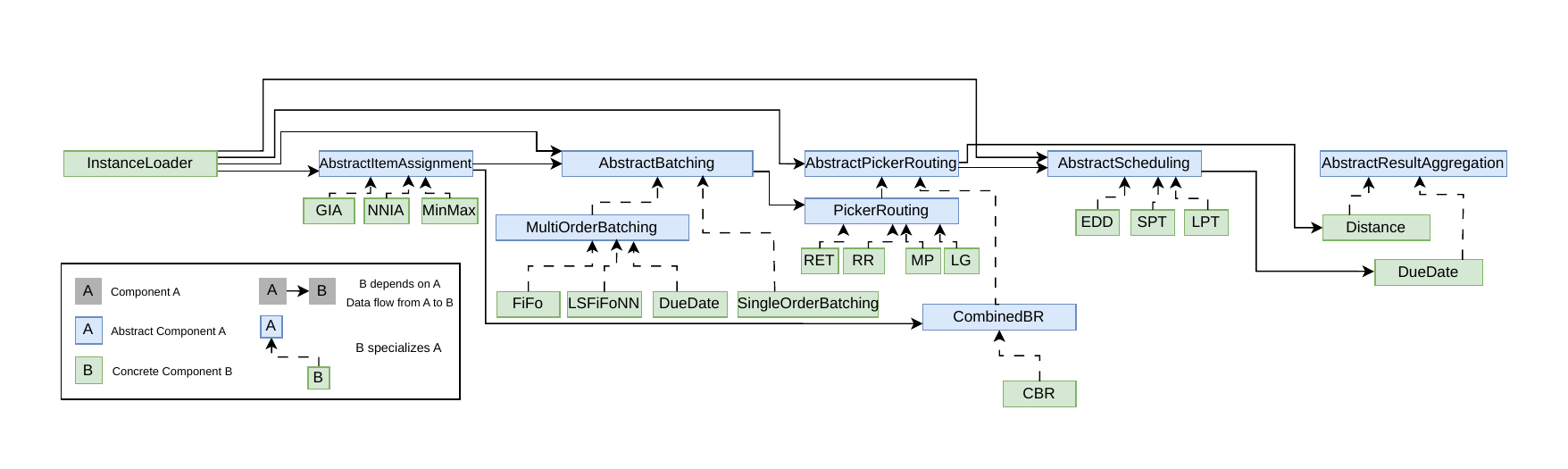}
\caption{Visualization of the \frameworkname pipeline template. For readability, not all concrete implementations are shown.}
\label{fig:template1}
\end{figure}

We implement the taxonomy from Figure~\ref{fig:taxonomy} and the data flow between subproblems as a CLS-Luigi pipeline template, shown in Figure~\ref{fig:template1}.


First, the \texttt{InstanceLoader} provides the domain objects and instance data required by the decision components. 
Further, problem specific components model the data flow of the problems defined in the taxonomy: 

\begin{itemize}
    \item \texttt{AbstractItemAssignment} produces resolved pick positions.

    \item \texttt{AbstractBatching} has two alternatives:
    \texttt{MultiOrderBatching}, which applies a batching algorithm, and \texttt{SingleOrderBatching}, which creates one batch per order. 

    \item \texttt{AbstractPickerRouting} produces routing solutions. It can be instantiated by
    standalone routing algorithms that consume a \texttt{BatchingSolution}, or by an integrated batching-and-routing
    component that depends directly on item assignment.

    \item \texttt{AbstractScheduling} consumes routing solutions and instance data. It assigns
    tours to resources and determines their execution sequence.
\end{itemize}

\texttt{ResultAggregation} is the terminal component used as synthesis target. 
It collects objective values, runtimes, and the algorithm configurations used in each executed pipeline.

The template distinguishes between abstract and concrete components. 
Abstract components represent the decision subproblems and define their typed inputs and outputs. 
Concrete components are the algorithm implementations that instantiate these abstract components. 

Algorithms from the repository described in Section~\ref{sec:algos} are added as concrete components. 
For example, batching algorithms instantiate \texttt{MultiOrderBatching}, routing algorithms instantiate the standalone routing alternative of \texttt{AbstractPickerRouting}, and scheduling approaches instantiate \texttt{AbstractScheduling}. 

Before synthesis, CASOP restricts the concrete components to the algorithms returned by the \texttt{DomainAlgoMapper}. 
Thus, only algorithms whose cards match the data card, problem class, objective, and feature constraints are available to CLS-Luigi.


Starting from \texttt{ResultAggregation}, CLS searches for all valid component combinations that can produce the requested result object. In our implementation, \texttt{ResultAggregation} is specialized into objective-specific targets such as \texttt{Distance} and \texttt{DueDate}.
Each valid inhabitant corresponds to one executable pipeline. 
The number of synthesized pipelines depends on the number of applicable algorithms per subproblem and on the structural alternatives encoded in the template.

\begin{figure}
    \centering
    \includegraphics[width=1\linewidth]{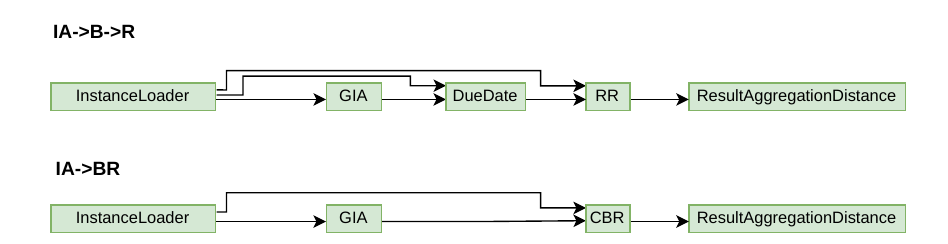}
    \caption{Example pipelines resulting from the defined template for the same problem (OBRP). The upper figure visualizes the OBRP in its decomposed, fully sequential form. The lower figure shows a partially integrated pipeline that combines batching and routing.}
    \label{fig:casop-non-lin}
\end{figure}

Figure~\ref{fig:casop-non-lin} shows two pipelines synthesized from the template for the OBRP problem class. 
The first follows the sequential decomposition: item assignment, batching, routing, and result aggregation. 
The second uses the integrated batching-and-routing component and therefore bypasses separate batching and standalone routing. 
Both pipelines are valid because they can produce the requested aggregation target.

After execution, \texttt{AbstractResultAggregation} collects the objective value, runtime, and concrete algorithm configuration of each executed pipeline. This is specialized into two variants for pipelines with and without time-related information.
CASOP then ranks all successfully solved pipelines according to the objective specified in the data card $\mathcal{I}$ and returns the best-performing pipeline.

\subsection{Pipeline Evaluator}
Based on the aggregated results, we identify the best-performing pipeline with respect to the defined objective $Z$, which is derived from the data card $\mathcal{I}$.
To this end, we apply a straightforward ranking mechanism that evaluates the output of \texttt{ResultAggregation} across all successfully solved pipelines and ranks them by objective value, thereby determining the overall best-performing pipeline.

\section{Framework Validation}
\label{sec:experiments}
We apply CASOP to seven benchmark data sets covering four problem types and warehouse settings that differ in layout, number of pickers, number of orders, storage policy, and available order information. The purpose of the experiments is to validate the implemented algorithms by comparing their solution quality to reported benchmark results, and answer the following three validation questions. 
\begin{itemize}
    \item[(Q1)] \textbf{Framework Adaptability:} Is the framework able to adapt to a wide range of warehouse and problem settings so that it can be adopted by practitioners?
    \item[(Q2)] \textbf{Algorithm Selection Potential:} Under what conditions does algorithm selection improve solution quality?
    \item[(Q3)] \textbf{Benchmarking:} How good is the performance of sequentially executed algorithms against the best known solutions? 
\end{itemize}
The remainder of this section is structured around these questions. 

All experiments were conducted on a Manjaro Linux 25.0.10 workstation with an Intel Xeon W5-2445 CPU (10 cores, 20 threads) and 125 GB RAM. 
CASOP is implemented in Python 3.13.7, and we use Gurobi 12.0.3 with default solver settings for the CBR model.
For the local search batching approach and the exact solution approaches, we set a time limit of 240 seconds, configurable through CLS-Luigi. 
All other heuristics were used without time limits. 
Each instance was processed by all applicable algorithm combinations.

\subsection{Validation Setup}
In the following, we briefly detail the validation setup.
Our validation applies CASOP to seven benchmark instances across four problem classes: SPRP, SPRP-SS, OBRP, and OBRSP. 
To demonstrate the broad applicability of CASOP, we collected representative instance sets from the literature covering a wide range of warehouse settings. The instance sets are characterized by varying levels of, e.g., pickers, capacities, storage policies, number of orders and orderlines, layout characteristics, and order-level information (due dates). 

\begin{table}[t]
  \centering
  \caption{Overview of instance sets used in the experimental framework validation.}
  \label{tab:problem_dimensions}
  \resizebox{\textwidth}{!}{
  \begin{tabular}{
    l l l
    r r r
    l r l
    r r r
    r
  }
    \toprule
    & & &
    \multicolumn{3}{c}{\textbf{Layout}} &
    \multicolumn{2}{c}{\textbf{Capacity}} &
    \multicolumn{2}{c}{\textbf{Storage}} &
    \multicolumn{2}{c}{\textbf{Orders}} &
    \multicolumn{1}{c}{\textbf{Resources}} \\
    \cmidrule(lr){4-6} \cmidrule(lr){7-8} \cmidrule(lr){9-10} \cmidrule(lr){11-12} \cmidrule(lr){13-13}
    \textbf{Problem} &
    \textbf{Instance Set} &
    \textbf{Reference} &
    \rotatebox{0}{\textbf{$n_b$}} & 
    \rotatebox{0}{\textbf{$n_a$}} & 
    \rotatebox{0}{\textbf{$n_p$}} & 
    \rotatebox{0}{\textbf{$\delta$}} & 
    \rotatebox{0}{\textbf{$c_k$}} & 
    \rotatebox{0}{\textbf{Storage $T$}} &
    \rotatebox{0}{\textbf{\# Policies}} &
    \rotatebox{0}{\textbf{$|\mathcal{O}|$}} & 
    \rotatebox{0}{\textbf{$|\mathcal{L}(o)|$}} & 
    \rotatebox{0}{\textbf{$|\mathcal{R}|$}} \\ 
    \midrule
    \multirow{2}{*}{SPRP}
    & \textit{SPRP}
    & \citet{hesler_exact_2024}
    & 1 & 5--50 & 30--180
    & --- & ---
    & Dedicated
    & 1
    & 1 & 3--30 & 1\\
    & \textit{SPRP-SS}
    & \citet{hesler_exact_2024}
    & 1 & 5--50 & 30--180
    & --- & ---
    & Scattered
    & 1
    & 1 & 3--30 & 1\\
    \midrule
    \multirow{4}{*}{OBRP}
    & \textit{BahceciOencan}
    & \citet{bahceci_evaluation_2022}
    & 1 & 10 & 10
    & \# Items & 5--30
    & Dedicated
    & 5
    & 8--12 & 10--29 & 1\\
    & \textit{HennWaescher}
    & \citet{henn_metaheuristics_2010}
    & 1 & 10 & 45
    & \# Items & 30--75
    & Dedicated
    & 2
    & 20--100 & 125--442 & 1\\
    & \textit{MuterOencan}
    & \citet{muter_exact_2015}
    & 1 & 10 & 10
    & \# Items & 24--48
    & Dedicated
    & 1
    & 20--100 & 62--200 & 1\\
    & \textit{Foodmart}
    & \citet{van_gils_formulating_2019}
    & 2 & 8 & 198
    & \# Items & 40
    & Dedicated
    & 1
    & 5--2000 & 59--8350 & 1\\
    \midrule
    OBRSP
    & \textit{Kris}
    & \citet{briant2023lowerupperboundsjoint}
    & 2 & 6--18 & 60--180
    & \# Orderlines & 4--45
    & Dedicated
    & 1
    & 6--300 & 1--45 & 2--6\\
    \bottomrule
    \multicolumn{12}{l}{\footnotesize Sources:
      \textsuperscript{a}\,\url{https://logistik.bwl.uni-mainz.de/forschung/benchmarks/}\,(SPRP, SPRP-SS, BahceciOencan, Henn, MuterOencan);} \\
    \multicolumn{12}{l}{\footnotesize \phantom{Sources:}
      \textsuperscript{b}\,\url{https://pagesperso.g-scop.grenoble-inp.fr/~cambazah/batching/}\,(Foodmart);
      \textsuperscript{c}\,\url{https://pagesperso.g-scop.grenoble-inp.fr/~cambazah/sequencing/}\,(Kris)} \\
  \end{tabular}
  }
\end{table}

\paragraph{Data Sources and Preprocessing}

We evaluate CASOP on seven benchmark instance sets covering different problem classes, warehouse layouts, and operational settings. An overview of all instance sets and their key characteristics is provided in Table~\ref{tab:problem_dimensions}.

We cover one SPRP instance set presented in \cite{hesler_exact_2024}. 
In the \textit{SPRP-SS} set, scattered storage is considered, where items may be stored in multiple storage locations. 
Both instance sets have a single picker and a single-block, parallel-aisle layout, with varying numbers of aisles and storage locations per aisle. 

For OBRP, we evaluate four instance sets. 
\textit{MuterOencan}, \textit{BahceciOencan}, and \textit{Henn} cover varying settings of single-block parallel-aisle layouts with varying picker capacities, number of orders, and number of articles per order. 
Each of these instance sets measures picker capacity in terms of the number of items. 
For \textit{BahceciOencan} and \textit{Henn}, five and two storage policies are varied, respectively. 
\textit{Foodmart} features a parallel-aisle layout with two blocks connected by a cross-aisle. 
For \textit{Foodmart}, picker capacity is constrained by the number of boxes on a picking cart. As boxes cannot contain mixed orders, each order $o \in O$ requires a dedicated number of boxes calculated as $b_o = \lceil \text{items}_o / B \rceil$, where $B$ denotes the box capacity. A batch is feasible if $\sum_{o \in b} b_o \leq K$, where $K$ is the number of boxes per cart as defined in the Resources domain $\mathcal{R}$. This constraint reflects the order indivisibility requirement from the original problem formulation by \cite{valle_optimally_2017}, where orders from different customers cannot share baskets.

The \textit{Kris} instance set covers the OBRSP and was first proposed in \cite{valle_optimally_2017} and further improved and published online by \cite{briant2023lowerupperboundsjoint}, who also publish solutions for a subset of 243 instances. 
\textit{Kris} includes multiple pickers, a two-block layout, and order due dates, with all orders being released at the beginning of the planning horizon. The \textit{Kris} instance set is provided with large and small instances that differ in the number of orders per instance, which we treat together. 

All instance sets originate from heterogeneous sources and formats. We therefore implement dedicated parsers and \texttt{DomainLoader} to transform each instance into the unified \frameworkname domain model. 
The preprocessing code and parsers are publicly available in our open-source repository to ensure reproducibility.

\paragraph{Employed Algorithms and Configurations}
We use a broad set of algorithms covering all subproblems defined in Section~\ref{sec:problem-setting}.
The employed methods and their abbreviations are summarized in Section \ref{sec:algos}. 

Our focus is on the combination of heuristic approaches, and thus, exact routing approaches are excluded from our evaluation. 
The only exceptions are the runtime efficient RR approach and CBR, which is included as an exact benchmark for routing and batching and to highlight the non-linear template structure. 
To keep experiment runtimes manageable, we evaluate CBR only on \textit{BahceciOencan}, the smallest OBRP instance set.

Batching methods that require configuration, e.g., with methods for initial solution construction, are evaluated using a restricted set of representative configurations.
\texttt{LocalSearchBatching} is combined with three routing algorithms (RR, NN, SShape) for solution evaluation and two initialization strategies (FiFo, RAND), yielding six LS configurations. 
\texttt{SeedBatching} is evaluated with the seed criterion \texttt{CLOSEST\_TO\_DEPOT} and two similarity measures (\texttt{SHARED\_ARTICLES}, \texttt{MIN\_DISTANCE}). For \texttt{SavingsBatching}, routing-based distance evaluation is performed using RR, NN, and SShape.

\paragraph{Performance Metrics and Objectives}
Following common practice in algorithm selection literature \citep{bischl2016aslib}, we compare the following:
\begin{itemize}
    \item Single Best Solver (SBS): The algorithm combination that achieves the best average performance across all instances in a instance sets.
    \item Virtual Best Solver (VBS): An oracle that always selects the best-performing algorithm for each instance. The VBS provides an upper bound on achievable performance through perfect algorithm selection.
    Because multiple strategies may yield the same result, we will use total runtime as a tiebreaker and always report the fastest strategy.
\end{itemize}
The relative gap between SBS and VBS quantifies the potential benefit of algorithm selection:
\[
\text{Relative Gap} = \frac{\text{SBS Mean} - \text{VBS Mean}}{\text{SBS Mean}} \times 100\%.
\]

We further report the gap between the best known solution (BKS) reported in literature and the VBS result per instance, i.e.,
\[
\text{Optimality Gap} = \frac{\text{VBS} - \text{BKS}}{\text{BKS}} \times 100\%.
\]

The runtime of the resulting pipelines is evaluated in seconds.  
In terms of objectives, we compare distance, makespan, tardiness, and on-time rate, based on the problem setting and instance features.

\subsection{Q1: Framework Adaptability Across Problem Classes}
We first analyze the pipeline generation process in general. We therefore evaluate the number of resulting pipelines per instance set and compare the runtimes of the three stages: instance generation, pipeline building, and execution. 

\paragraph{Pipeline Overview}
Table~\ref{tab:pipeline_results} summarizes the execution of CASOP, reporting the number of synthesized pipelines per instance set alongside the number of applicable algorithm components for item assignment, batching, routing, and scheduling.
The results demonstrate how the filtering mechanism identifies applicable algorithms based on problem type and instance characteristics, effectively pruning the combinatorial space of possible pipelines.
We denote the filtered algorithm configurations per subproblem by $\mathit{IA}^* \subseteq \mathit{IA}$, $\mathit{B}^* \subseteq \mathit{B}$, $\mathit{R}^* \subseteq \mathit{R}$, $\mathit{BR}^* \subseteq \mathit{BR}$, and $\mathit{S}^* \subseteq \mathit{S}$.

For the \textit{SPRP} and \textit{SPRP-SS} instance sets, which consist of single-order instances without batching capacity, no genuine order-batching decision is present.
For consistency with the pipeline template, the table reports one single-order batching alternative, which acts as an identity component.
The resulting number of pipelines per instance is therefore $|\mathit{IA}| \times |\mathit{B}| \times |\mathit{R}|$, with $|\mathit{B}|=1$.

For OBRP instance sets, batching algorithms are applicable as well. For \textit{BahceciOencan}, the BR subproblem is applicable.
Therefore, the resulting number of pipelines for this instance set is not only the Cartesian product $|\mathit{IA}^*| \times |\mathit{B}^*| \times |\mathit{R}^*|$, but also includes the pipelines generated by the combined batching-and-routing approach, $|\mathit{IA}^*| \times |\mathit{BR}^*|$.

For this instance set, the framework synthesizes both distance-based pipelines without scheduling and time-based pipelines with scheduling.
The resulting number of pipelines per instance is therefore
\[
|\mathit{IA}^*| \times |\mathit{B}^*| \times |\mathit{R}^*|
+
|\mathit{IA}^*| \times |\mathit{B}^*| \times |\mathit{R}^*| \times |\mathit{S}^*|.
\]

These results demonstrate that CASOP is able to capture the domain features and algorithm requirements and successfully synthesizes valid algorithm pipelines for four different problem classes based on this information with a single CLS-Luigi template. 

\begin{table}
\centering
\caption{Number of resulting pipelines per instance set.}
\label{tab:pipeline_results}
\begin{tabular}{lrrrrrrr}
\toprule
  & \multicolumn{5}{c}{\# Algorithms} & \# Instances & \# Pipelines\\ 
\cmidrule(lr){2-6}
Instance Set & $\mathit{IA}$ & $\mathit{B}$ & $\mathit{R}$ & $\mathit{BR}$ & $\mathit{S}$ &  &  \\
\midrule
\textit{SPRP} & 1 & 1 & 6 & 0 & 0 & 2400 & 14400 \\
\textit{SPRP-SS} & 5 & 1 & 6 & 0 & 0 & 14300 & 429000 \\
\textit{BahceciOencan} & 1 & 11 & 6 & 1 & 0 & 1350 & 90450 \\
\textit{HennWaescher} & 1 & 11 & 6 & 0 & 0 & 5759 & 380094 \\
\textit{MuterOencan} & 1 & 11 & 6 & 0 & 0 & 270 & 17820 \\
\textit{FoodmartData} & 1 & 9 & 5 & 0 & 0 & 144 & 6480 \\
\textit{Kris} & 1 & 13 & 5 & 0 & 3 & 480 & 124800 \\
\bottomrule
$\sum$ &  &  &  &  &  & 24,704 & 1,063,044\\
\bottomrule
\end{tabular}
\end{table}

\paragraph{Runtimes}
Table \ref{tab:runtimes-detailed} reports the computational overhead of the pipeline synthesis process.
\begin{table}[ht]
\centering
\caption{Detailed pipeline generation runtimes (mean $\pm$ std, in seconds)}
\label{tab:runtimes-detailed}
\begin{tabular}{l r r r r}
\toprule
Instance Set & Load Domain & Build Pipelines & Run Pipelines & Total \\
\midrule
\textit{SPRP} & $1.1717 \pm 2.7287$ & $0.221 \pm 0.189$ & $0.388 \pm 0.722$ & $1.781 \pm 3.464$ \\
\textit{SPRP-SS} & $1.4288 \pm 3.3660$ & $0.569 \pm 0.332$ & $0.465 \pm 0.835$ & $2.463 \pm 4.202$ \\
\textit{BahceciOencan} & $0.0088 \pm 0.0083$ & $0.878 \pm 0.125$ & $154.078 \pm 109.904$ & $154.965 \pm 109.906$ \\
\textit{HennWaescher} & $0.0224 \pm 0.0276$ & $0.354 \pm 0.167$ & $25.956 \pm 30.621$ & $26.333 \pm 30.590$ \\
\textit{MuterOencan} & $0.0025 \pm 0.0004$ & $0.235 \pm 0.011$ & $27.489 \pm 24.348$ & $27.728 \pm 24.347$ \\
\textit{Foodmart} & $0.0851 \pm 0.0431$ & $0.190 \pm 0.100$ & $19.728 \pm 53.565$ & $20.003 \pm 53.576$ \\
\textit{Kris} & $0.4311 \pm 0.5602$ & $23.146 \pm 22.500$ & $184.554 \pm 297.259$ & $208.132 \pm 284.392$ \\
\bottomrule
\end{tabular}
\end{table}
The runtimes are given for the initial domain generation (load domain), the pipeline build process (build pipelines), the sequential execution of the resulting pipelines (run pipelines), and the total time (total).  
Notably, the load domain and build pipelines phases remain negligible across all instance sets, except for \textit{Kris}, confirming that the framework's overhead for algorithm filtering and pipeline construction is minimal compared to actual algorithm execution.
For \textit{SPRP-SS}, the entire process completes in under 3 seconds per instance on average. 
Across all instances, the total runtime is dominated by the run pipelines phase, which executes all synthesized algorithm pipelines using Luigi. 

The long pipeline runtimes of \textit{Kris} instances indicate that the time limits for the LS batching configurations are exhausted, and the relative complexity of the batching phase compared to other instance sets, where solutions can be found faster. 
This can also be seen in \textit{BahceciOencan}, where the longer runtimes are attributable to the combined batching-and-routing approach.
Also, for the other instance sets, high standard deviation indicates instance-dependent variability, likely driven by varying instance sizes and greater computational effort, e.g., in generating the distance matrix and graph. At the same time, this serves as a good example of the caching mechanism of CLS-Luigi. When running new algorithms for the same instances, these time-consuming preprocessing steps do not have to be repeated.

To summarize Q1, the experiments demonstrate that CASOP successfully processes instances across multiple problem classes, varying layout representations, storage policies, and resource configurations. The mapping approach implemented in CASOP automatically identifies applicable algorithms based on warehouse characteristics and algorithm requirements. Through direct comparison with the best reported results from the literature, we could also verify the correctness of the instance modeling and algorithm implementations.  

\subsection{Q2: Algorithm Selection Potential}
Table \ref{tab:vbs_overview_all} shows the comparison of VBS vs. SBS for all instance sets aggregated by the applicable objectives. 
\begin{table}[!h]
\centering
\caption{VBS overview.}
\label{tab:vbs_overview_all}
\begin{adjustbox}{width=\linewidth,center}
\begin{tabular}{lllrrrl}
\toprule
Problem & Instance Set & SBS Strategy & SBS Mean & VBS Mean & Relative Gap \% & Objective \\
\midrule
SPRP & \textit{SPRP} & GIA+RawInput+RR & 589.83 & 589.83 & 0.00 & distance\\
SPRP & \textit{SPRP-SS} & MinMinIA+RawInput+RR & 395.28 & 385.78 & 2.40 & distance \\
OBRP & \textit{BahceciOencan} & GIA+CBR & 168.28 & 168.10 & 0.11 & distance\\
OBRP & \textit{HennWaescher} & GIA+SavingsRR+RR & 8114.34 & 8083.95 & 0.37 & distance \\
OBRP & \textit{MuterOencan} & GIA+SavingsRR+RR & 1293.23 & 1279.49 & 1.06 & distance \\
OBRP & \textit{Foodmart} & GIA+SavingsNN+NN & 2299.58 & 2095.33 & 8.88 & distance \\
OBRSP & \textit{Kris} & GIA+SavingsNN+NN+EDD & 144076.33 & 143695.60 & 0.26 & distance \\
OBRSP & \textit{Kris} & GIA+SavingsNN+NN+LPT & 71368.43 & 70044.13 & 1.86  & makespan \\
OBRSP & \textit{Kris} & GIA+DueDate+NN+EDD & 2153.53 & 2037.06 & 5.41 & max\_tardiness \\
OBRSP & \textit{Kris} & GIA+DueDate+NN+EDD & 96.49 & 97.94 & 1.50 & on\_time\_rate \\
\bottomrule
\end{tabular}

\end{adjustbox}
\end{table}
We report the name of the SBS as a concatenation of the pipeline's algorithmic components in the order of item assignment, pick list generation, picker routing, and scheduling.
We further report the mean distance of the SBS (SBS mean), the mean distance of the VBS (VBS mean), and the gain achieved by the VBS as relative gap between SBS and VBS (Relative Gap \%).  
For most instance sets, the potential gain expected from the VBS is marginal and largely depends on the applicable solution approaches. 
The largest gains can be observed for the \textit{Foodmart} and \textit{SPRP-SS} instance sets, with a gain of 8.88 \% and 2.40 \%, respectively. 
For other instance sets such as \textit{SPRP}, \textit{Kris}, and \textit{BahceciOencan}, we find a gain of around 0 \% for the distance objective. This indicates that there is a single strategy that performs best on every instance of the respective instance set. The reason in the case of \textit{SPRP}, and \textit{BahceciOencan} is the use of approaches that provide exact results.

For \textit{HennWaescher} and \textit{MuterOencan}, CBR was not used, but we can still see that the combination of SavingsRR and RR wins on almost every instance, which again is not surprising, as both approaches use the exact RR routing approach. The LS configurations were not able to outperform the savings approach within the specified time limit. 

\begin{figure}
    \centering
    \includegraphics[width=1\linewidth]{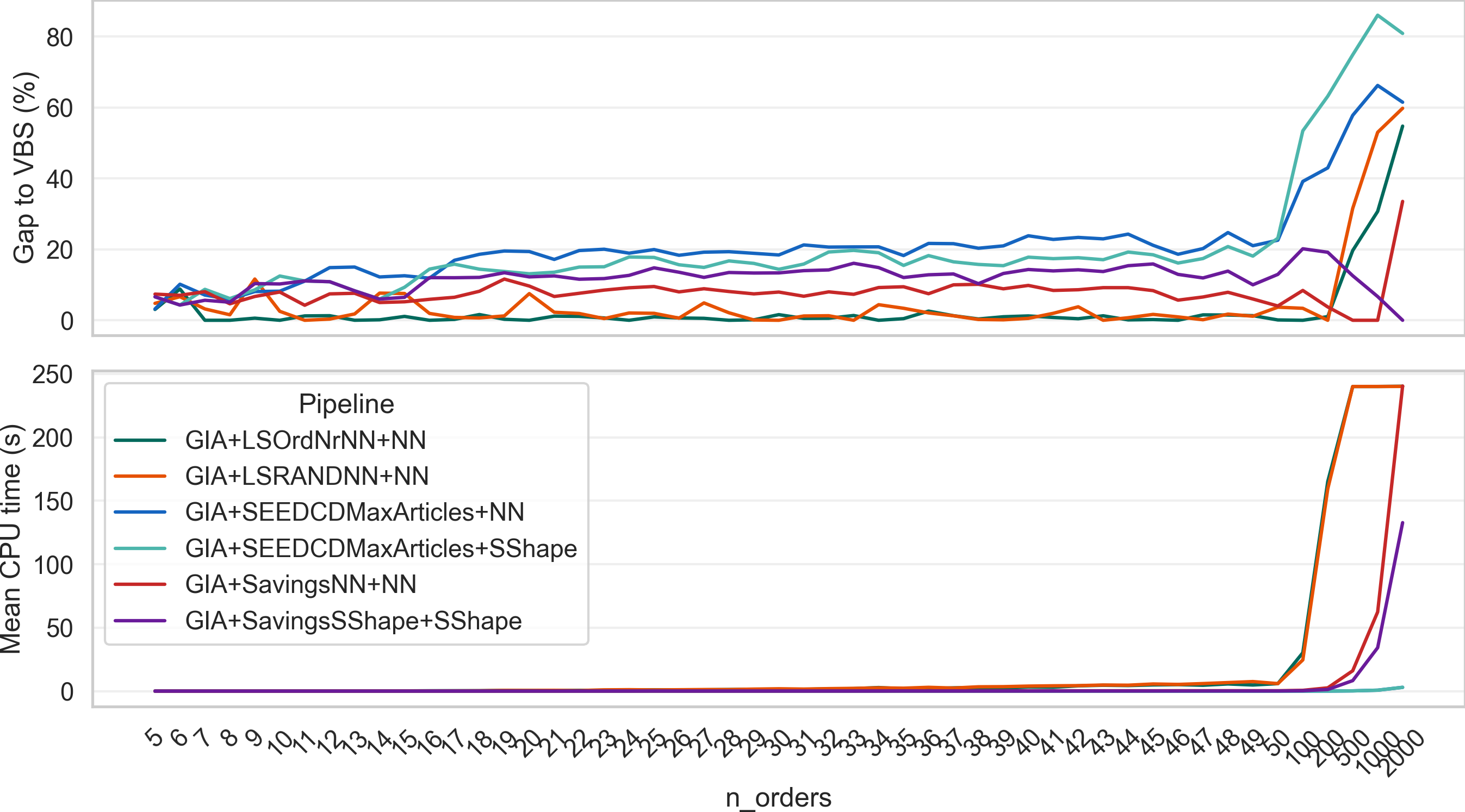}
    \caption{Per-strategy gap to the VBS (top) and mean CPU time (bottom) on the \textit{Foodmart} instance set, grouped by number of orders.}
    \label{fig:foodmart_gap}
\end{figure}

\textit{Foodmart} has the largest relative gap between SBS and VBS. To investigate this, we analyze the per-instance gap between each strategy and the VBS as a function of the number of orders. Figure \ref{fig:foodmart_gap} shows the gap to the VBS (top) and the mean CPU time (bottom) for all strategies that achieve the best result on at least one instance. For small and medium instances (up to 100 orders), the local search configurations GIA+LSOrdNrNN+NN and GIA+LSRANDNN+NN consistently achieve the lowest gaps. However, for instances with 500 or more orders, their CPU time approaches the 240-second time limit, and the gap increases to over 80\%. 
The savings- and seed-based strategies maintain stable gaps across all instance sizes at low computational cost. The local search configurations dominate on small instances but exhaust the 240-second time limit on large instances, degrading their mean performance. As the SBS is determined by the overall mean, it falls on GIA+SavingsNN+NN. Figure \ref{fig:foodmart_gap} indicates that  this strategy actually performs worse than the LS based strategy for all instances below 200 orders. This also highlights the drawback of the pure mean-based determination of the SBS. 

\begin{figure}
    \centering
    \includegraphics[width=1\linewidth]{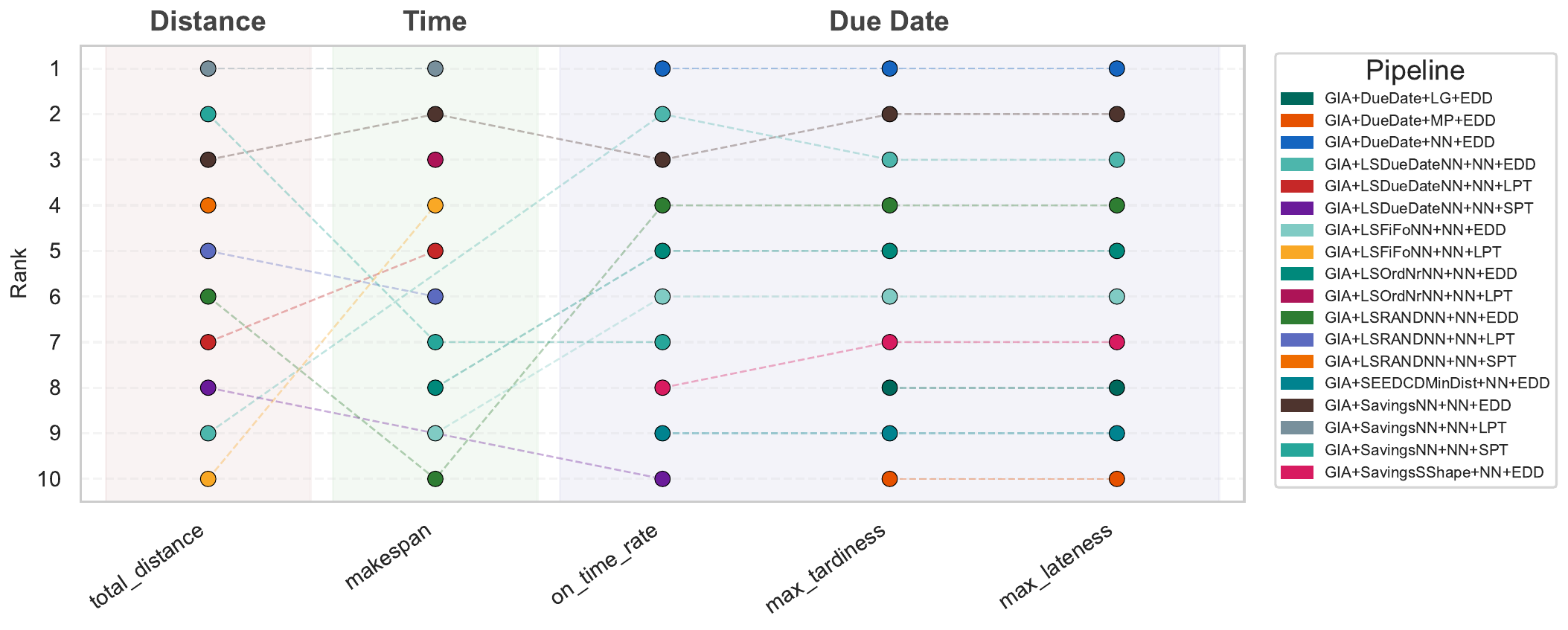}
    \caption{Varying ranks of strategies for different objectives on the \textit{Kris} instance set. Results are grouped by objective and objective categories. We show the ten best performing strategies for the respective objective, ranked from best to worst.}
    \label{fig:kris_ranking}
\end{figure}

Since the \textit{Kris} instances allow us to solve the scheduling subproblem, we can analyze time-related objectives.  
The comparison of the SBS reveals that for different objectives, different strategies may perform better, and rankings across different objectives are not consistent, as illustrated in Figure \ref{fig:kris_ranking}. 
For the distance and makespan objectives, the pipeline GIA+SavingsNN+NN+LPT performs best. 
However his pipeline is not in the top ten for the due-date-related objectives, where GIA+DueDate+NN+EDD consistently performs best. 
This is because the pipelines for distance and makespan ignore the due-date constraints and can therefore find solutions that are better than those that sort by due date. 

To summarize Q2, algorithm selection improves solution quality when the best-performing pipeline depends on instance characteristics or on the objective function. For most distance-based objectives, the relative gap between SBS and VBS is small, indicating that a single robust strategy often dominates across the entire instance set. This is especially visible for \textit{SPRP}, \textit{BahceciOencan}, \textit{HennWaescher}, \textit{MuterOencan}, and \textit{Kris} under the distance objective. However, the \textit{Foodmart} results clearly show the potential of algorithm selection: local-search configurations perform best on small and medium instances, whereas savings- and seed-based strategies are more stable on larger instances where local search runs out of time. The \textit{Kris} results further indicate that the best pipeline depends on the objective, since distance, makespan, tardiness, and on-time rate favor different strategy rankings. Algorithm selection is therefore most useful in heterogeneous settings where instance size, time limits, or objective functions change the relative performance of the available pipelines.

\subsection{Q3: Benchmarking Against Best Known Solutions}

We compare the VBS results to the BKS values from the corresponding benchmark studies, where possible. These results contain both heuristic and optimal solutions. For \textit{Kris}, the objective values are calculated from the reported solution schedules.

Table \ref{tab:validation_summary} reports the number of benchmark instances (\# Instances), the evaluated objective (Objective), the mean relative gap of the VBS to the BKS (Mean Gap to BKS [\%]), the ex-post runtime of the selected VBS pipeline (VBS Runtime [s]), and the corresponding mean objective values of the VBS and the BKS (VBS Mean and BKS Mean).
\begin{table}[!ht]
\centering
\caption{Validation results across instance sets.}
\label{tab:validation_summary}
\begin{adjustbox}{width=\linewidth,center}
\begin{tabular}{lrlrrrr}
\toprule
Instance Set & \# Instances & Objective & Mean Gap to BKS [\%] & VBS Runtime [s] & VBS Mean & BKS Mean\\
\midrule
\textit{SPRP} & 2400 & distance & 0.000 & 0.006 & 589.834 & 589.834 \\
\textit{SPRP-SS} & 14300 & distance & 9.574 & 0.004 & 385.784 & 345.356 \\
\textit{BahceciOencan} & 1350 & distance & 0.015 & 36.333 & 168.102 & 168.074 \\
\textit{HennWaescher} & 1440 & distance & 3.206 & 4.085 & 8090.171 & 7840.128 \\
\textit{MuterOencan} & 270 & distance & 2.458 & 7.144 & 1308.775 & 1278.174 \\
\textit{Foodmart} & 42 & distance & 6.296 & 0.307 & 850.801 & 799.302 \\
\textit{Kris} & 294 & distance & 6.655 & 0.067 & 69828.082 & 64261.571 \\
\textit{Kris} & 294 & on-time rate & 0.945 & 0.072 & 99.050 & 99.993 \\
\textit{Kris} & 294 & on-time\_distance & 13.945 & 0.062 & 81898.231 & 64261.571 \\
\bottomrule
\end{tabular}
\end{adjustbox}
\end{table}

\begin{figure}[!h]
    \centering
    \includegraphics[width=1\linewidth]{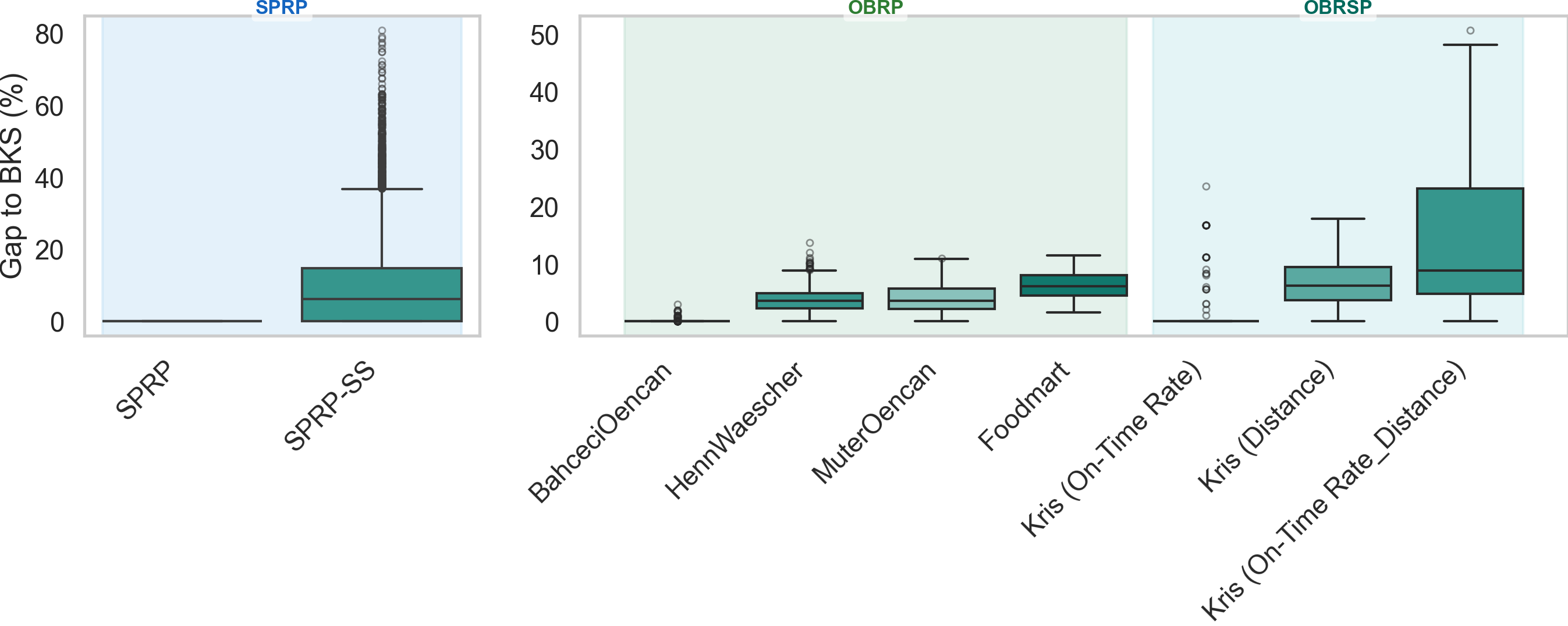}
    \caption{Comparison of gap between VBS and BKS on different instance sets. Results are grouped by instance set and problem class and sorted in ascending order by average gap of the VBS. While the instance sets for the problem classes SPRP and OBRP show the distance objective, for \textit{Kris}, three objectives are evaluated, which are specified in parentheses.}
    \label{fig:vbs_gap_to_lit}
\end{figure}

Figure \ref{fig:vbs_gap_to_lit} visualizes the gap between the VBS and the optimal results as boxplots.

As expected, the framework achieves the best results for instance sets with parallel-aisle single-block layouts (\textit{SPRP}, \textit{BahceciOencan}, \textit{MuterOencan}), where exact approaches for the subproblems are applicable. 
For \textit{SPRP}, we can validate the correctness of our RR implementation and our instance representation, as it matches the reported results in \cite{hesler_exact_2024}. The VBS reaches a Mean Gap to BKS of 0.000\%, with identical VBS Mean and BKS Mean values of 589.834.

For \textit{SPRP-SS}, the remaining gap is mainly caused by the fact that the item assignment is solved with a heuristic component, as opposed to the integrated approach presented in \cite{hesler_exact_2024}.
On the \textit{BahceciOencan} instance set, the average gap to the best reported results is close to zero, indicating that the integrated CBR model is able to reproduce the optimal results reported in \cite{hesler_modeling_2022}.
The results of \textit{MuterOnecan} and \textit{HennWaescher}, for which CBR was not used, show that heuristic batching approaches combined with RR routing are strong baselines, achieving Mean Gap to BKS values of 2.458\% and 3.206\%, respectively.  
For \textit{Foodmart}, we observe a Mean Gap to BKS of 6.296\%. This instance set uses a two-block layout, making RR routing inapplicable to the evaluated implementation. 

For the \textit{Kris} instance set, we evaluate three objectives: on-time rate, distance objective, and the hierarchical objective of on-time rate and distance. 
For the hierarchical objective, the VBS selects the pipeline with the highest on-time rate per instance, using distance only as a tiebreaker.
This is the fairest comparison, as using the VBS solely for the distance objective would not respect the due-date constraints the literature model is subject to. 
Although the VBS achieves a Mean Gap to BKS below 1\% for the on-time rate objective, this does not translate to the hierarchical objective where the Mean Gap to BKS is 13.945\%.
When simply selecting the pipeline with the shortest distance, we can achieve a Mean Gap to BKS of 6.655\%. 

To summarize Q3, the benchmark comparison shows that the VBS can reproduce or closely approach best reported solution values when the evaluated component portfolio matches the benchmark setting. For \textit{SPRP}, the RR implementation exactly matches the reported results, validating both the routing implementation and the instance representation. For \textit{BahceciOencan}, the integrated CBR model yields a Mean Gap to BKS close to zero.
For settings without CBR, the results quantify the performance of sequential batching-and-routing pipelines. The VBS achieves an average gap of 3.206\% for \textit{HennWaescher} and 2.458\% for \textit{MuterOencan}. Larger gaps occur, for instance, in sets where the evaluated portfolio relies on heuristic components or where specific exact components are not applicable or not yet implemented in the algorithm repository, as in \textit{SPRP-SS}, \textit{Foodmart}, and the hierarchical objective on \textit{Kris}.

\section{Conclusion and Future Research}
\label{sec:conclusion}
This paper introduced the CASOP framework for context-aware synthesis of optimization pipelines in warehouse optimization. Building on the need to study decomposed planning models in warehouse operations, the framework addresses how order fulfillment problems can be decomposed into subproblems, how applicable algorithm configurations can be identified from warehouse context and algorithm requirements, and how these configurations can be composed into executable pipelines.

CASOP combines four main components. First, warehouse systems are described using structured, machine-readable data cards that capture the problem class, objective, and available warehouse data across the five domain objects: layout, articles, orders, resources, and storage. Second, optimization algorithms are represented as algorithm cards that specify the addressed subproblem, optimization objective, and requirements regarding domain types, features, and feature constraints. Third, a context-aware mapping procedure matches algorithm requirements against the data card and the problem taxonomy to identify algorithms applicable to a given warehouse setting. Fourth, the resulting algorithms are composed into executable pipelines using CLS-Luigi. The taxonomy is translated into a pipeline template that covers both sequential decompositions and partially integrated solution paths, such as integrated batching and routing, and is instantiated using algorithms from the open-source repository provided with the framework, which includes implementations for item assignment, order batching, picker routing, integrated batching and routing, and picker scheduling.

The experimental evaluation shows that CASOP can be applied across heterogeneous warehouse optimization settings. Across seven benchmark instance sets covering four problem classes, the framework generated and evaluated 1,063,044 valid pipelines on 24,704 instances. The comparison with reported benchmark results further shows the performance range of the evaluated component portfolio. For \textit{BahceciOencan}, the integrated CBR model yields an average gap close to zero, while sequential heuristic pipelines achieve average gaps of 3.43\% and 4.85\% for \textit{HennWaescher} and \textit{MuterOencan}, respectively. Larger gaps occur in settings where the evaluated portfolio relies on heuristic components because exact-algorithm components are not applicable, for example, in scattered-storage instances or two-block layouts.

Overall, the paper demonstrates that context-aware pipeline synthesis is a feasible approach for structuring and evaluating decomposed decision-making approaches in warehouse optimization. Instead of studying isolated algorithms or manually selected combinations, the proposed framework makes the relation between warehouse context, algorithm applicability, pipeline structure, and resulting performance explicit.

Based on this, several directions for future work emerge. 
First, while this paper focuses on order fulfillment, the methodology can be transferred to other warehouse optimization problems and potentially to broader supply chain planning settings.
Second, the algorithm repository should be extended to integrate additional subproblems and algorithm variants for existing ones. 
Third, future work should study dynamic and online warehouse settings, where the applicable pipeline may change as orders, resources, and system states evolve.
Finally, recent developments in large language models and agentic systems may support selected parts of the framework, such as generating algorithm cards, refining pipeline templates, or integrating new algorithm components.



\section*{Acknowledgements}
This publication was produced as part of the research project ``Data- and target-driven sequential decision-making for time-dynamic logistics systems'', which was funded by the Deutsche Forschungsgemeinschaft (DFG, German Research Foundation) -- 502552827.




\bibliographystyle{elsarticle-harv} 
\bibliography{references}

\end{document}